\documentclass[10pt,journal,compsoc]{IEEEtran}
\ifCLASSOPTIONcompsoc
  \usepackage[nocompress]{cite}
\else
  \usepackage{cite}
\fi
\ifCLASSINFOpdf
  \usepackage[pdftex]{graphicx}
\else
\fi

\usepackage{hyperref}

\usepackage[utf8]{inputenc}
\usepackage[T1]{fontenc}
\usepackage{subfigure} 
\usepackage{multirow}
\usepackage[english]{babel}
\usepackage[usenames, dvipsnames]{color}
\usepackage{color}
\usepackage{xcolor}
\usepackage{standalone}
\usepackage{tikz}

\DeclareMathSymbol{\varTheta}{\mathord}{letters}{"02}
\usepackage{rotating}
\hypersetup{draft}

\begin{document}

\title{Survey on Emotional Body Gesture Recognition}

\author{Fatemeh~Noroozi,
        Ciprian~Adrian~Corneanu,
        Dorota~Kamińska,
        Tomasz~Sapiński,
        Sergio~Escalera,
        and~Gholamreza~Anbarjafari,% <-this % stops a space
\IEEEcompsocitemizethanks{\IEEEcompsocthanksitem F. Noroozi and G. Anbarjafari are with the iCV Research Group, Institute of Technology, University of Tartu, Tartu, Estonia.\protect\\
E-mail: \{fatemeh.noroozi,shb\}@ut.ee
\IEEEcompsocthanksitem D. Kamińska and T. Sapiński are with Department of Mechatronics, Lodz University of Technology, Lodz, Poland.\protect\\
E-mail: dorota.kaminska@p.lodz.pl, sapinski.tomasz@gmail.com
\IEEEcompsocthanksitem C. A. Corneanu and S. Escalera are with the University of Barcelona and Computer Vision Center, Barcelona, Spain.\protect\\
E-mail: cipriancorneanu@gmail.com, sergio@maia.ub.es
\IEEEcompsocthanksitem G.~Anbarjafari is also with Department of Electrical and Electronic Engineering, Hasan Kalyoncu University, Gaziantep, Turkey.%
}% <-this % stops a space
\thanks{Manuscript received January 19, 2018; revised Xxxx XX, 201X.}}

\markboth{Journal of IEEE Transactions on Affective Computing,~Vol.~XX, No.~X, Xxx~201X}%
{Noroozi \MakeLowercase{\textit{et al.}}: Survey on Emotional Body Gesture Recognition}

\IEEEtitleabstractindextext{%
\begin{abstract}
Automatic emotion recognition has become a trending research topic in the past decade. While works based on facial expressions or speech abound recognizing affect from body gestures remains a less explored topic. We present a new comprehensive survey hoping to boost research in the field. We first introduce emotional body gestures as a component of what is commonly known as "body language" and comment general aspects as gender differences and culture dependence. We then define a complete framework for automatic emotional body gesture recognition. We introduce person detection and comment static and dynamic body pose estimation methods both in RGB and 3D. We then comment the recent literature related to representation learning and emotion recognition from images of emotionally expressive gestures. We also discuss multi-modal approaches that combine speech or face with body gestures for improved emotion recognition. While pre-processing methodologies (e.g. human detection and pose estimation) are nowadays mature technologies fully developed for robust large scale analysis, we show that for emotion recognition the quantity of labelled data is scarce, there is no agreement on clearly defined output spaces and the representations are shallow and largely based on naive geometrical representations.    
\end{abstract}

}

\maketitle

% is not selected <OR> if conference mode is selected - because compsoc
% conference papers position the abstract like regular (non-compsoc)
\IEEEdisplaynontitleabstractindextext

% \fi
\IEEEpeerreviewmaketitle
%%%*********************** Introduction***********************************
\section{Introduction}
\label{sec:intro}
\begin{figure}[t]
\centering
\includegraphics[width=0.45\textwidth]{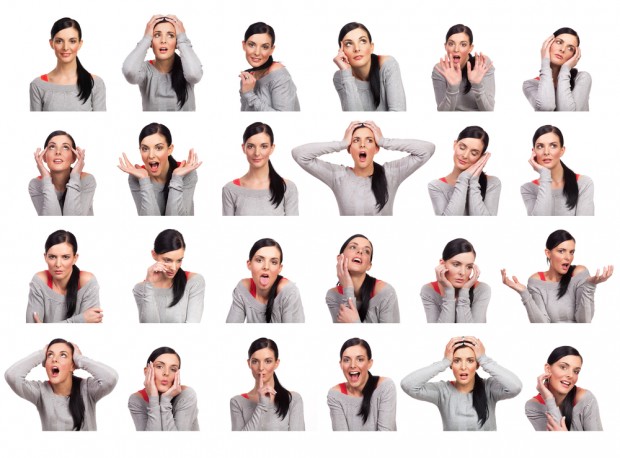}
\caption{Body language includes different types of nonverbal indicators such as facial expressions, body posture, gestures and eye movements. These are important markers of the emotional and cognitive inner state of a person. In this work, we review the literature on automatic recognition of body expressions of emotion, a subset of body language that focuses on gestures and posture of the human body. The images have been taken from~\cite{smrtic2015asertivna}.}
\label{fig:body_language}
\end{figure}

\IEEEPARstart{D} {uring} conversations people are constantly changing nonverbal clues, communicated through  body movement and  facial expressions. The difference between the words people pronounce and our understanding of their content comes from nonverbal communication also commonly called body language. Some examples of body gestures and postures, key components of body language are shown in Fig. \ref{fig:body_language}.

Although it is a significant aspect of human social psychology, the first modern studies concerning body language has become popular in 1960s \cite{pease}. Probably the most important work published before 20th century was \textit{The Expression of the Emotions in Man and Animals} by Charles Darwin \cite{darwin1998expression}. This work is the foundation of modern approach to body language and many of Darwin's observations were confirmed by subsequent studies. Darwin observed that people all over the world use facial expressions in a fairly similar manner. Following this observation, Paul Ekman researched patterns of facial behavior among different cultures of the world. In 1978, Ekman and Friesen developed the Facial Action Coding System (FACS) to model human facial expressions \cite{ekman}. In an updated form, this descriptive anatomical model is still being used in emotion expressions recognition.

The study of use of body language for emotion recognition was conducted by Ray Birdwhistell who found that the final message of an utterance is affected only 35\% by the actual words and 65\% by non-verbal signals \cite{bird}. In the same work, analysis of thousands of negotiations recordings revealed that the body language decides the outcome of those negotiations in 60\% - 80\% of cases. The research also showed that during a phone negotiation, stronger arguments win, however during a personal meeting, decisions are made on the basis of what we see rather than what we hear \cite{pease}.
At the present, most researchers agree that words serve primarily to convey information and  the body movements to form relationships and sometimes even to substitute the verbal communication (e.g. lethal look). 

Gestures are one of the most important forms of nonverbal communication. They include movements of hands, head and other parts of the body that allow individuals to communicate a variety of feelings, thoughts and emotions. Most of the basic gestures are the same all over the world: when we are happy we smile when we are upset we frown~\cite{pease2008definitive}, \cite{rosenstein1988differential}, \cite{ekman1987universals}. 

According to~\cite{pease}, gestures can be of the following types:
\begin{itemize}
  \item Intrinsic: For example, nodding as a sign of affirmation or consent is probably innate, because even people who are blind from birth use it;
  \item Extrinsic: For example, turning to the sides as a sign of refusal is a gesture we learn during early childhood - It happens when, for example, a baby has had enough milk from the mother's breast, or with older children when they refuse a spoon during feeding;
  \item A result of natural selection: For example, the expansion of the nostrils to oxygenate the body can be mentioned, which takes place when preparing for battle or escape.
\end{itemize}
The ability to recognize the attitude and thoughts from one's behavior was the original system of communication before the speech. Understanding of emotional state enhances the interaction. Although computers are now a part of human life, the relation between a human and a machine is not natural. Knowledge of the emotional state of the user would allow the machine to adapt better and generally improve cooperation. 

While emotions can be expressed in different ways, automatic recognition has mainly focused on facial expressions and speech.  About 95\% of the literature dedicated to this topic focused on faces as a source for emotion analysis \cite{gelder}. Considerably less works were done on body gestures and posture. With recent developments of motion capture technologies and reliability, the literature about automatic recognition of expressive movements grew significantly.

Despite the increasing interest in this topic, we are aware of just a few relevant survey papers. For example, Kleinsmith et al. \cite{survey1} reviewed the literature on affective body expression perception and recognition with an  emphasis on inter-individual differences, impact of culture and multi-modal recognition. In another paper, Kara et al. \cite{survey2} introduced categorization of movement into four types: communicative,  functional, artistic, and abstract and discussed the literature associated with these types of movements.

In this work, we cover all the recent advancements in automatic emotion recognition from body gestures. The reader interested in emotion recognition from facial expressions or speech is encouraged to consult dedicated surveys~\cite{corneanu2016survey,escalera2017challenges,asadi2017deep}. In this work we refer to these only marginally and only as complements to emotional body gestures. In Sec. \ref{sec:expression} we briefly introduce key aspects of affect expression through body language in general and we discuss in-depth cultural and gender dependency. Then, we define a standard pipeline for automatically recognizing body gestures of emotion in Sec. \ref{sec:recognition} and we discuss in details technical aspects of each component of such pipeline. Furthermore, in Sec. \ref{sec:data} we provide a comprehensive review of publicly available databases for training such automatic recognition systems. 
%In Sec. \ref{sec:multimodality} we discuss recognition of affect in a complementary way form facial expressions, speech and body gestures. 
We conclude in Sec. \ref{sec:discussion} with discussions and potential future lines of research.

%[SURVEY] Kleinsmith, Andrea, and Nadia Bianchi-Berthouze. "Affective body expression perception and recognition: A survey." IEEE Transactions on Affective Computing 4.1 (2013): 15-33.

%[SURVEY] Karg, Michelle, et al. "Body movements for affective expression: A survey of automatic recognition and generation." IEEE Transactions on Affective Computing 4.4 (2013): 341-359.
 
%%%*********************** Expression***********************************

\section{Expressing emotion through  body language}
\label{sec:expression}
According to~\cite{iaccino2014left,ruthrof2015body} body language includes different kinds of nonverbal indicators such as facial expressions, body posture, gestures, eye movement, touch and the use of personal space. The inner state of a person is expressed through elements such as iris extension, gaze direction, position of hands and legs, the style of sitting, walking, standing or lying, body posture, and movement. Some examples are presented in Fig. \ref{fig:body_language}.

\iffalse
\begin{figure}[t]
\centering
\includegraphics[width=0.48\textwidth]{face-body.jpg}
\caption{Sample facial expressions (a1–h1) and associated body postures (a2–h2): Neutral (a), Negative surprise (b), Positive surprise (c), Boredom (d), Uncertainty (e), Anxiety (f-g) and Puzzlement (h)~\cite{gunes2009automatic}.}
\label{F14}
\end{figure}
\fi 

After the face, hands  are probably the richest source of body language information~\cite{molchanov2015hand,pease2016definitive}. For example, based on the position of hands one is able to determine whether a person is honest (one will turn the hands inside towards the interlocutor) or insincere (hiding hands behind the back). Exercising open-handed gestures during conversation can give the impression of a more reliable person. It is a trick often used in debates and political discussions. It is proven that people using open-handed gestures are perceived positively \cite{pease}. 
%For example, a study of two groups of speakers in public, showed that the ones using open gestures were perceived positively by 85\% of their recipients, whereas those who had their palms facing downwards were evaluated as positive only by 52\% of the receivers \cite{pease}. Moreover, the position of fingers is of great importance when interpreting gestures. For example, the index finger pointed outwards, with all other fingers clenched is associated negatively (as a threatening expression). The same speakers received only 28\% of positive feedback after using this gesture during the presentation.
%Most people are aware of what their face and hands can tell about their internal state. When they try to control the body language, they focus mainly on their facial expressions and hands gestures. People are least aware of the position of their legs, and in particular the feet. Paul Ekman noted that during lying or under stress one radically increases the number of unconscious foot movements (jiggling). The legs can reveal if a person is open and comfortable (one foot positioned slightly forward), or if they are negative and uncertain (crossed legs). Position of the legs during sitting is also an important source of information. For example, if one sits with slightly open legs (relaxed position), it means the person is comfortable, on the other hand knees held gently or tightly together indicate anxiety. The very way of walking also tells a lot about personality (e.g. military march aims to show strength and vitality). 

Head positioning also reveals a lot of information about emotional state. The research ~\cite{pease2008definitive} indicates that people are prone to talk more if the listener encourages them by nodding. The pace of the nodding can signal patience or lack of it. In neutral position the head remains still in front of the interlocutor. If the chin is lifted it may mean that the person is displaying superiority or even arrogance. Exposing the neck might be a signal of submission. In \cite{darwin1998expression} Karl Darwin noted that like animals, people tilt their heads when they are interested in something. That is why women perform this gesture when they are interested in men, an additional display of submission results in greater interest from the opposite sex, 
e.g. a lowered chin signals a negative or aggressive attitude.

The torso is probably the least communicative part of the body. However, its angle with the body is an indicative attitude. For example placing the torso frontally to the interlocutor can be considered as a display of aggression. By turning it at a slight angle one may be considered self-confident and devoid of aggression. Leaning forward, especially when combined with nodding and smiling, is the most distinct way to show curiosity \cite{pease2008definitive}.

The above considerations indicate that in order to correctly interpret body language as indicators of emotional state, various parts of body must be considered at the same time. According to~\cite{siegman2014nonverbal}, body language recognition systems may benefit from a variety of psychological behavioral protocols. An example of general movements protocol for six basic emotions is presented in Table~\ref{T111}.
\begin{table*}[t]
	\centering
	\caption{The general movement protocols for the six basic emotions \cite{gunes2005fusing, gunes2006bimodal,gunes2015bodily}.}
	\label{T111}
	\tiny
	\resizebox{\textwidth}{!}{
	    \begin{tabular}{|c|p{.5\textwidth}|} \hline
	    \textbf{Emotion} & \textbf{Associated body language} \\ \hline\hline
	    Fear & Noticeably high heart beat-rate (visible on the neck). Legs and arms crossing and moving.  Muscle tension: Hands or arms clenched, elbows dragged inward, bouncy movements, legs wrapped around objects. Breath held. Conservative body posture. Hyper-arousal body language. \\ \hline
	    Anger & Body spread. Hands on hips or waist. Closed hands or clenched fists. Palm-down posture. Lift the right or left hand up. Finger point with right or left hand. Finger or hand shaky. Arms crossing. \\ \hline
	    Sadness & Body dropped. Shrunk body. Bowed shoulders. Body shifted. Trunk leaning forward. The face covered with two hands. Self-touch (disbelief), body parts covered or arms around the body or shoulders. Body extended and hands over the head. Hands kept lower than their normal positions, hands closed or moving slowly. Two hands touching the head and moving slowly. One hand touching the neck. Hands closed together. Head bent. \\ \hline
	    Surprise & Abrupt backward movement. One hand or both of them moving toward the head. Moving one hand up. Both of the hands touching the head. One of the hands or both touching the face or mouth. Both of the hands over the head. One hand touching the face. Self-touch or both of the hands covering the cheeks or mouth. Head shaking. Body shift or backing. \\ \hline
	    Happiness & Arms open. Arms move. Legs open. Legs parallel. Legs may be stretched apart. Feet pointing something or someone of interest. Looking around. Eye contact relaxed and lengthened. \\ \hline
	    Disgust & Backing. Hands covering the neck. One hand on the mouth. One hand up. Hands close to the body. Body shifted. Orientation changed or moving to a side. Hands covering the head. \\ \hline
		\end{tabular}
	}
\end{table*}

\subsection{Culture differences}

\iffalse
\begin{figure}[t]
\centering
\includegraphics[width=0.48\textwidth]{cooper.png}
\caption{Classification of different countries based on the willingness of their people to touch one another in exchanging greetings. The chart has been taken from~\cite{westsidetoastmasters}.}
\label{fig:cooper}
\end{figure}
\fi 

It has been reported that gestures are strongly culture-dependent \cite{efron1941gesture,kendon1983study}. However, due to exposure to mass-media, there is a tendency of globalization of some gestures especially in younger generations \cite{pease2008definitive}. This is despite the fact that the same postures might have been used for expressing significantly different feelings by their previous generations. Consequently, over time, some body postures might change in meaning, or even disappear. For instance, the thumb-up symbol might have different meanings in different cultures. In Europe it stands for number "1" in Japan for "5", while in Australia and Greece, using it may be considered insulting. However, nowadays, it is widely used as a sign of agreement, consent or interest \cite{westsidetoastmasters}.

Facial expressions of emotion are similar across many cultures \cite{ekman1992argument}. This might hold in the case of postures as well. In \cite{camras1992japanese}, the effect of culture and media on emotional expressions was studied. One of the conclusions was that an American and a Japanese infant present closely similar emotional expressions. Most of the studies reported on this topic in the literature inferred that intrinsic body language, gestures and postures are visibly similar throughout the world. However, a decisive conclusion still requires more in-depth exploration, which is challenging due to the variety of topics that need to be studied on numerous cultures and countries. Therefore, the researchers investigating this issue prefer to concentrate on a certain activity, and study it on various cultures, which may lead to a more understandable distinction. For example, in many cultures, holding hands resembles mutual respect, but in some others touching one another in exchanging greetings might not be considered usual \cite{westsidetoastmasters}. %The results from the foregoing study are shown in Fig. \ref{fig:cooper}.

\subsection{Gender differences}

\begin{figure}
    \centering
    \includegraphics[width=\linewidth, height=4.5cm]{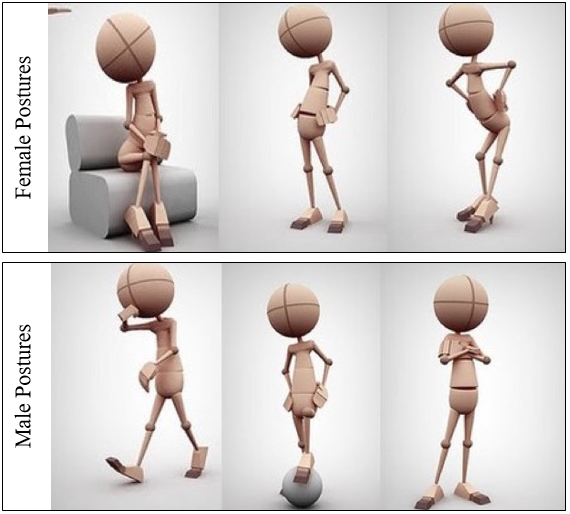}
    \caption{There are fundamental differences in the way men and women communicate through body language. In certain situations, one can easily discriminate the gender when only the body pose is shown. Illustration from \cite{gender_diffs}.}
    \label{fig:gender_diffs}
\end{figure}

Women are believed to be more perceptive than men due to the concept of female intuition \cite{stand_out}.
%During a research study carried out at Harvard University, scientists asked group of men  And women to decode a short movie with no audio signal. 87\% of women and 42\% of an men aptly specified the course of the conversation. This observation is confirmed by brain tomography. There are more areas responsible for behavioral evaluation in the female brain then in the male brain \cite{gender0}. Several studies have shown that both males and females differ significantly in identifying affect \cite{gender1}, \cite{gender2}. In \cite{gender3}, it is mentioned that women are better at detecting facial expressions than men. The authors proved that there are no significant differences in recognition accuracy when labeling explicit expressions, but there is  a significant difference in recognizing subtle emotions. In \cite{gender4}, basing on detailed multimodal research, the authors confirmed that women are more accurate in the recognition of emotional prosody. A further review \cite{gender8} confirmed female advantage in emotion recognition when investigating infants, children and adolescents. What is more, Simon Baron-Cohen \cite{gender5} proved that women much better recognize a person's mental state solely on the basis of the eyes.
There are some fundamental differences in the way women and man communicate through body language \cite{gender_diffs} (see Fig. \ref{fig:gender_diffs} for some trivial examples). 
This may be caused by influence of culture (tasks and expectations that face both sexes), body composition, makeup and worn type of clothes. 
%In \cite{gender7} researchers  found evidence for gender differences in emotion expression investigating children and teenagers. They proved that girls express more happiness, sadness, anxiety, and shame or embarrassment than boys, while boys express more anger and are more prone to externalize emotions, such as contempt. These studies show that clear differences can already be seen in childhood.

Women wearing mini skirts often sit with crossed legs or ankles. But it is not the sole reason of this gesture applying  almost exclusively to women. As a result of body composition, most men are not able to sit that way, thus this position became a symbol of femininity. Another good example is the cowboy pose popularized by old western movies. In this pose the thumbs are placed in the belt loops or pockets with the remaining fingers pointed downwards towards the crotch. Men use this gesture when they defend their territory or while demonstrating their courage. A similar position has been also observed among monkeys \cite{pease}.

Generally, women show emotions and their feelings more willingly than men \cite{simon2004gender}, which are associated with qualities such as kindness, supportiveness, affection and care for others. Men are more likely to display power and dominance while simultaneously hiding the melting mood \cite{simon2004gender}. However, nowadays these general tendencies start to faint and are considered as gender stereotypes \cite{hess2000emotional}. 

%%%*********************** Modelling***********************************
\section{Models of the human body and emotion}
\label{sec:modelling}
Before discussing the main steps for automatically recognizing emotion from body gestures, (details in Sec. \ref{sec:recognition}), we first discuss modelling of the input and output of such systems. The input will be an abstraction of the human body (and its dynamics) that we would like to map through machine learning methods to a certain predefined abstraction of emotion. Deciding the appropriate way of modelling the human body and emotion is an essential design decision. We  begin by discussing abstractions of the human body (Sec. \ref{sec:modelling:model_body}) and then main models of emotion used in affective computing (Sec. \ref{sec:modelling:model_emotion}). 

\subsection{Models of the human body}
\label{sec:modelling:model_body}

\begin{figure}
    \centering
    \includegraphics[width=\linewidth]{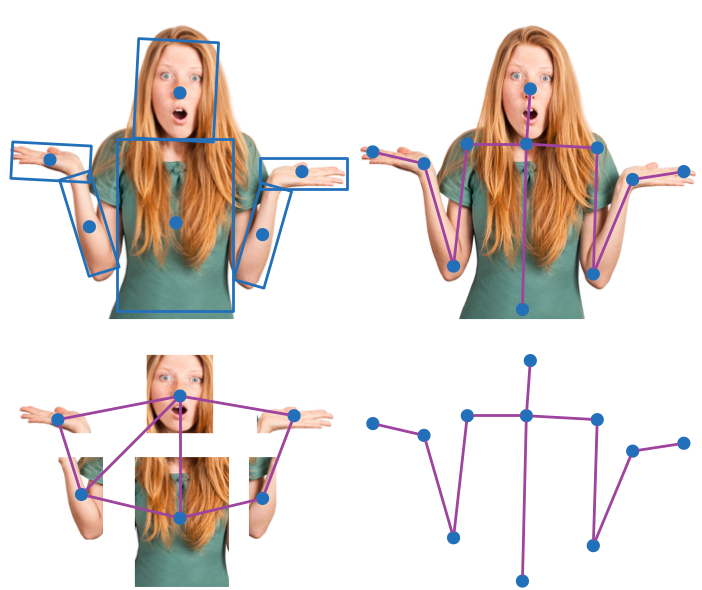}
    \caption{The two most common ways of modelling the human body in automatic processing are either as an ensemble of body parts or as a kinematic model. In ensemble of body parts (left) different parts of the body are independently detected and soft restrictions can be imposed to refine these detections. A kinematic model (right) is a collections of interconnected joints with predefined degrees of freedom similar to the human skeleton.}
    \label{fig:human_models}
\end{figure}

Human body has evolved such that it can perform complex actions, which require coordination of various organs. Therefore, many everyday actions present unique spatio-temporal movement structures~\cite{bernhardt2010emotion}. In addition, some pairs of body actions, e.g. walking and drinking, may be performed at the same time and their expression might not be additive ~\cite{wu2007joint}. The two main lines for abstracting the human body have been following either a constrained composition of human body parts or a kinematic logic based on the skeletal structure of the human body (see Fig. \ref{fig:human_models}).   

\iffalse
It has been argued that recognizing the emotion from everyday actions demands detecting the type of the action~\cite{bernhardt2010emotion}.
For example, in case the purpose of a hand raise is knocking on a door, the speed of the motion and the height of the lift can indicate the emotional state of the subject~\cite{paterson2001role} and the aim of knocking, respectively. However, if the purpose is to gently lift a glass of water, the speed and the height are both constrained by the action.

Human gait \cite{whittle2014gait} is the most studied type of action and the most energy-efficient form of locomotion \cite{bernhardt2010emotion}. It can be affected by gender~\cite{lee2002gait}, long-term elements, e.g. height~\cite{benabdelkader2002stride}, mid-term parameters, including age and body weight~\cite{everett2010human}, and short-term factors, such as emotion~\cite{crane2007motion,janssen2008recognition}. Therefore, human gait can be investigated in order to detect the foregoing factors, including emotions. In that case, analyzing finer specifications of the orchestration of the movement components is required for recognizing parameters other than the broad action category, e.g. understanding that the subject is angry, on top of recognizing the action as knocking on a door. For example, after recognizing an action as walking, other features, such as stride length and cadence~\cite{crane2007motion,benabdelkader2002stride}, should be analyzed in order to extract other factors.
\fi

\begin{figure}[t]
\centering
\includegraphics[width=0.48\textwidth]{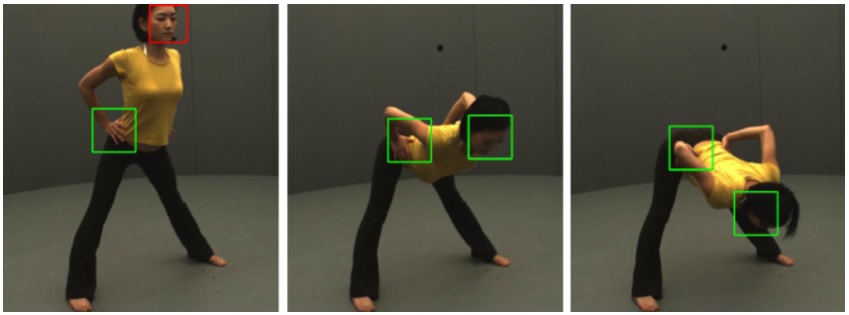}
\caption{Example of body pose estimation and tracking using ensemble of parts namely, head and hands ~\cite{tung2008human}.}
\label{fig:pictorial_structures}
\end{figure}

\textbf{Part Based Models.}
In a part based approach the human body is represented as flexible configuration of body parts. Body parts can be detected independently (face, hands, torso) and priors can be imposed using domain knowledge of the human body structure to refine such detection. Some examples of ensemble of parts models of the human body are pictorial structures and grammar models. Pictorial structures are generative 2D assemblies of parts, where each part is detected with its specific detector. Pictorial structures are a general framework for object detection widely used for people detection and human pose estimation \cite{fischler1973representation}. An example of body pose estimation using pictorial structures is shown in Fig. \ref{fig:pictorial_structures}.

Grammar models provide a flexible framework for detecting objects, which was also applied for human detection in \cite{felzenszwalb2011object}. Compositional rules are used to represent objects as a combination of other objects. In this way, human body could be represented as a composition of trunk, limbs and face; as well composed by eyes, nose and mouth.

\begin{figure*}[t]
\centering
\includegraphics[width=\textwidth]{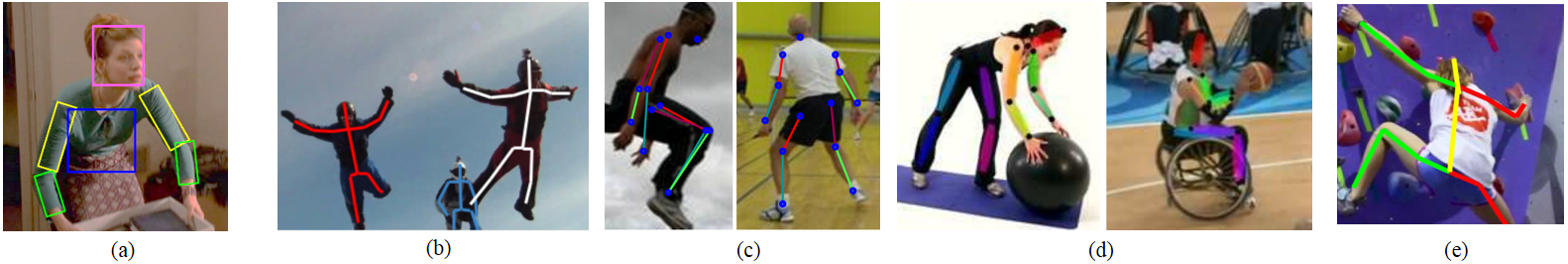}
\caption{Examples of body pose estimation using artificial neural networks: (a)  latent structured support
vector machines LSSVM ~\cite{zhang2014latent}, (b) Associative embedding supervised convolutional neural network (CNN) for group detection~\cite{newell2017associative}, (c) Hybrid architecture consisting of a deep CNN and a Markov Random Field ~\cite{tompson2014joint}, (d) Replenishing back-propagated gradients and conditioning the CNN procedure~\cite{wei2016convolutional} and (e) CNN by using the iterative error
feedback processing ~\cite{carreira2016human}.}
\label{fig:kinematic_deep}
\end{figure*}

\textbf{Kinematic Models.}
Another way of modelling the human body is by defining a collection of interconnected joints also known as kinematic chain models. This is usually a simplification of the human skeleton and its mechanics. A common mathematical representation of such models is through a cyclical tree graphs which also present the advantage of being computationally convenient. Contrary to part based approach \cite{fischler1973representation}, nodes of structure trees represent joints, each one parameterized with its degrees of freedom. Kinematic models can be planar, in which case they are a projection in the image plane or depth information can be considered as well. Richer, more realistic variants can be defined for example as a collection of connected cylinders or spheroids or 3D meshes. Examples of body pose detection using kinematic models and deep learning methods are shown in Fig. \ref{fig:kinematic_deep}.

\subsection{Models of emotion}
\label{sec:modelling:model_emotion}
\begin{figure}
    \centering
    \includegraphics[width=\linewidth]{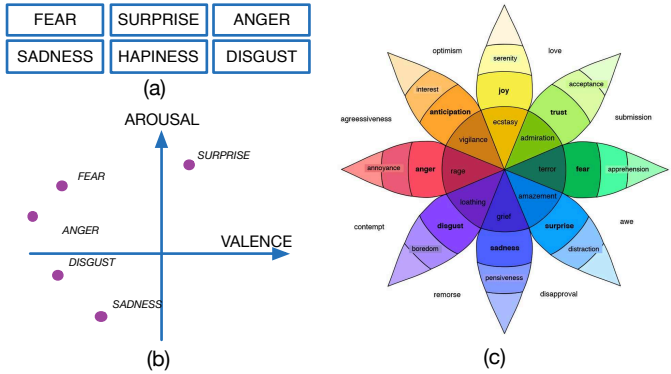}
    \caption{There are three main ways of modelling emotion in affective computing. Emotion can be defined (a) in a categorical way (here a universal set of emotions as define by Ekman \cite{ekman1971universal}) (b) as point in space defined along a set of latent dimensions (Russells model depicted) \cite{russell1977evidence} or (c) in a hybrid manner (Plutchik model shown) \cite{plutchik2001nature}.}
    \label{fig:emotion_models}
\end{figure}

The best way of modelling affect has been subject of debate for a long time and many perspectives upon the topic were proposed. The most influential models (and in general most relevant for affective computing applications) can be classified in three main categories: categorical, dimensional and componential \cite{kolakowska2015modeling} (see Fig. \ref{fig:emotion_models} for examples of each category). 

\textbf{Categorical models.} Classifying emotions into a set of distinct classes that can be recognized and described easily in daily language has been common since at least the time of Darwin. More recently, influenced by the research of Paul Ekman \cite{ekman1971universal, ekman1994strong} a dominant view upon affect is based on the underlying assumption that humans universally express and recognize a set of discrete primary emotions which include happiness, sadness, fear, anger, disgust, and surprise. Mainly because of its simplicity and its universality claim, the universal primary emotions hypothesis has been by far the first choice for affective computing research and has been extensively exploited.

\textbf{Dimensional models.} Another popular approach is to model emotions along a set of latent dimensions \cite{greenwald1989affective, russell1977evidence, watson1988development}. These dimensions include valence (how pleasant or unpleasant a feeling is) activation (how likely is the person to take action under the emotional state) and control (the sense of control over the emotion). Due to their continuous nature, such models can theoretically describe more complex and subtle emotions. Unfortunately, the richness of the space is more difficult to use for automatic recognition systems because it can be challenging to link such described emotion to a body expression of affect. This is why, many automatic systems based on dimensional representation of emotion simplified the problem by dividing the space in a limited set of categories like positive vs negative or quadrants of the 2D space \cite{zeng2009survey}.\\ 
\textbf{Componential models.} Somehow in-between categorical and dimensional models in terms of descriptive capacity, componential models of affect, arrange emotions in a hierarchical fashion where each superior layer contains more complex emotions which can be composed of emotions of previous layers . The best example of componential models was proposed by Plutchik \cite{plutchik2001nature}. According to his theory, more complex emotions are combinations of pairs of more basic emotions, called dyads. For example, love is considered to be a combination of joy and trust. Primary dyads, e.g. optimism=anticipation+joy, are often felt, secondary dyads, e.g. guilt=joy+fear, are sometimes felt and tertiary dyads, e.g. delight=joy+surprise, are seldom felt. These types of models are rarely used in affective computing literature compared to the previous two but should be taken into consideration due to their effective compromise between ease of interpretation and expressive capacity. 

%%%*********************** Recognition***********************************
%\section{Automatically recognizing body gestures of emotion}
\section{Body Gesture Based Emotion Recognition}
\label{sec:recognition}

\begin{figure*}[t]
\centering
\includegraphics[width=\textwidth, height=6cm]{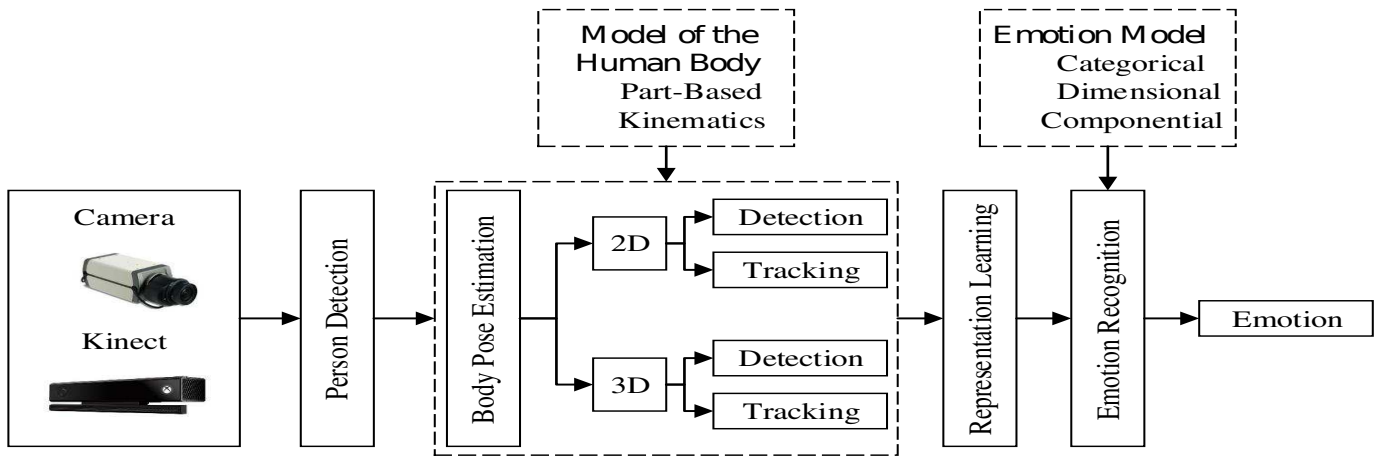}
\caption{General overview of an Emotion Body Gesture Recognition system. After detecting persons in the input for background extraction, a common step is to estimate the body pose. This is done either by detecting and tracking different parts of the body (hands, head, torso, etc.) or by mapping a kinematic model (a skeleton) to the image. Based on the extracted model of the human body, a relevant representation is extracted or learned in order to map the input to a predefined emotion model using automatic pattern recognition methods.}
\label{Overview}
\end{figure*}

 In this section, we present the main components of what we call an Emotion Body Gesture Recognition (EBGR) system. For a detailed depiction see Fig. \ref{Overview}. 
An important preparation step, which influences all the subsequent design decisions for such an automatic pipeline is the determination of the appropriate modelling of input (human body) and targets (emotion). Depending on the type of the model that has been chosen, either a publicly accessible database can be utilized, or a new one needs to be created. Similarly, other elements of the system need to be selected and configured such that they are compatible with each other, and overall, provide an efficient performance. Regardless of the foregoing differences between various types of EBGR systems, the common first step is to detect the body as a whole, i.e. to subtract the background from every frame which represents a human presenting a gesture. We will briefly discuss the main literature for human detection in Sec. \ref{sec:human_detection}. The second step is detection and tracking of the human pose in order to reduce irrelevant variation of data caused by posture (we dedicate Sec. \ref{sec:tracking} to this). The final part of the pipeline, which we discuss in Sec. \ref{sec:representation_and_recognition}, consists in building an appropriate representation of the data and applying a learning technique (usually classification or regression) to map this representation to the targets. We conclude this section with a presentation of the most important applications of automatic recognition of emotion using body gesture.

\subsection{Human Detection}
\label{sec:human_detection}
Human detection in images usually consists in determining rectangular bounding boxes that enclose humans. It can be a challenging task because of the non-rigid nature of the human body, pose and clothing, which result in high variation of appearance. In uncontrolled environments changes of illumination and occlusions add to the complexity of the problem. 

A human detection pipeline follows the general pipeline of object detection problems and consists of extracting potential candidate regions, representing those regions, classifying the regions as human or non-human, and merging positives into final decisions \cite{nguyen2016human}. If depth information is available, it can be used to  limit the search space and considerably simplify the background substraction problem \cite{nguyen2016human}. Modern techniques might not exactly follow this modularization, either by jointly learning representation and classification or by directly proposing detection regions from input.  

One of the first relevant methods for human detection was propose by Viola and Jones~\cite{viola2003detecting}. Following a method previously applied to face detection, it employs a cascade structure for efficient detection, and utilizing AdaBoost for automatic feature selection \cite{viola2003detecting}. 

An important advancement in performance came with the adoption of gradient-based features for describing shape. Dalal and Triggs, popularized the so called histogram of oriented gradient (HOG) features for object detection by showing substantial gains over intensity based features \cite{dalal2006human}. Since their introduction, the number of variants of HOG features has proliferated greatly with nearly all modern detectors utilizing them in some form \cite{dollar2012pedestrian}.

Earlier works on human detection assumed no prior knowledge over the structure of the human body. Arguably one of the most important contributions in this direction was the Deformable Part Models (DPM) \cite{felzenszwalb2008discriminatively}. A DPM is a set of parts and connections between the parts which relate to a geometry prior. In the initial proposal by Felzenswalb et al. a discriminative part based approach models unknown part positions as latent variables in a support vector machine (SVM) framework. Local appearance is easier to model than global appearance and training data can be shared across deformations. Some authors argued that there is still no clear evidence for the necessity of components and parts, beyond the case of occlusion handling \cite{benenson2014ten}.

In late years a spectacular rise in performance in many pattern recognition problems was brought by training deep neural networks (DNN) with massive amounts of data. According to some authors, the obtained results for human detection are on par with classical approaches like DPM, making the advantage of using such architectures yet unclear \cite{benenson2014ten}. Evenmore, DNN models are known to be very slow, especially when used as sliding-window classifiers which makes it challenging to use for pedestrian detection. One way to alleviate this problem is to use several networks in a cascaded fashion. For example, a smaller, almost shallow network was trained to greatly reduce the initially large number of candidate regions produced by the sliding window. Then in a second step, only high confidence regions were passed through a deep network obtaining in this way a trade-off between speed and accuracy \cite{angelova2015real}. The idea of cascading any kind of features of different complexity, including deeply learnt features was addressed by seeking an algorithm for optimal cascade learning under a criterion that penalizes both detection errors and complexity. This made it possible to define quantities such as complexity margins and complexity losses, and account for these in the learning process. This algorithm was shown to select inexpensive features in the early cascade stages, pushing the more expensive ones to the later stages \cite{cai2015learning}. State-of-the-art accuracy with large gains in speed were reported when detecting pedestrians.

For a comprehensive survey of the human detection literature, the interested reader is referred to \cite{nguyen2016human} and \cite{dollar2012pedestrian,benenson2014ten}.

\subsection{Body Pose Detection}
\label{sec:tracking}
Once background has been subtracted, detecting and tracking the human body pose is the second stage in automatic recognition of body gestures of affect. This consists in estimating the parameters of a human body model from a frame or from a sequence of frames, while the position and configuration of the body may change~\cite{mikic2003human}.

\subsubsection{Body Pose Detection}
Due to the high dimensions of the search space and the large number of degrees of freedom, as well as variations of cluttered background, body parameters and illumination, human pose estimation is a challenging task~\cite{kar2010skeletal,anbarjafari2015video}. It also demands avoiding body part penetration and impossible positions.

Body pose estimation can be performed using either model fitting or learning. Model-based methods fit an expected model to the captured data, as an inverse kinematic problem~\cite{barron2000estimating,taylor2000reconstruction}. In this context, the parameters can be estimated based on the tracked feature points, using gradient space similarity matching and the maximum likelihood resulted from the Markov Chain Monte Carlo approach~\cite{siddiqui2010human}. However, model-based methods are not robust against local extrema, and require initialization~\cite{kar2010skeletal}.

%On the other hand, 
Performing pose estimation using learning is computationally expensive, because of the high dimensions of the data, and requires a large database of labeled skeletal data. Using poselets for encoding the pose information was proposed in~\cite{bourdev2009poselets,gall2009motion}. They utilized SVM for classification of the results of skeletal tracking.

Using parallelism for clustering the appearances was proposed in~\cite{ramanan2003finding}. Hash-initialized skeletons were tracked through range images using the Iterative Closest Point~(ICP) algorithm in~\cite{grest2005nonlinear}. Segmentation and classification of the vertices in a closed 3D mesh into different parts for the purpose of human body tracking was proposed in~\cite{kalogerakis2010learning}.

Haar-cascade classifiers were trained for the segmentation of head and upper-body parts in~\cite{kar2010skeletal}. They proposed a hybrid approach which involves model-based fitting as well. The results of the learning and classification procedures were combined with the information from extended distance transform and skin segmentation, in order to fit the data to the skeletal model.

Starting with the DeepPose~\cite{toshev2014deeppose}, the field of Human Pose Estimation (HPE) experienced a considerable change from using traditional approaches to developing based on deep networks. In the aforementioned study, the 2D joint coordinates were directly regressed using a deep network. In~\cite{tompson2014joint}, heatmaps were created based on multiple resolutions of a given image at the same time, in order to obtain the features according to numerous scales.

In fact, utilizing CNNs for 3D HPE~\cite{lin2017recurrent,coskun2016human,li20143d,chen20163d,mehta2017monocular} is meant to tackle the limitedness of the applicability of classical methods. The latter usually can be trained only based on the few existing 3D pose databases. In~\cite{tome2017lifting}, the probabilistic 3D pose data were fused by using a multi-stage CNN, which were then considered in order to refine the 2D locations, according to the projected belief maps.

The input and model for a deep learning procedure can be of various types. For example, in~\cite{newell2016stacked}, simple nearest neighbor upsampling and multiple bottom-up, top-down inferences through stacking numerous hourglasses were proposed, as well as combining a Convolutional Network (ConvNet) with a part-based spatial model. The foregoing approach was computationally efficient as well, which helps take advantage of higher spatial precisions~\cite{tompson2014joint}.

Dual-source CNN (DS-CNN)~\cite{fan2015combining} is another framework which can be utilized in the context of HPE. Example results obtained by utilizing the foregoing strategy from different studies are shown in Fig.~\ref{fig:kinematic_deep}.

In~\cite{newell2016stacked}, each body part was first individually analyzed by the network, which resulted in a heatmap and the corresponding associative embedding tag. After comparing the joint embedding tags, separate pose predictions were made~\cite{mehta2017monocular}. Heatmaps were taken into account in order to find a 2D bounding box locating the subject using a CNN referred to as 2DPoseNet. Afterward, another CNN, namely, 3DPoseNet, regressed the 3D pose, followed by calculating the global 3D pose and perspective correction based on the camera parameters.

During the last few years, numerous studies have utilized deep learning methods for feature extraction. For example, in~\cite{toshev2014deeppose,newell2016stacked,yang2016end}, seven-layered convolutional DNNs were considered for regressing and representing the joint contexts and locating the body. In~\cite{ouyang2014multi}, high-level global features were obtained from different sources, and then combined through a deep model. In~\cite{tompson2014joint}, a deep convolutional network was combined with the Markov random field, in order to develop a part-based body detector according to multi-scale features.

In~\cite{neverova2013multi,metallinou2013tracking}, each gesture was modeled based on combinations of large-scale motions of body parts, such as torso, head or limbs, and subtle movements, e.g. those of fingers. The data gathered from the whole time-span were then combined using recursive neural networks (RNN) and long short-term memory (LSTM). Similarly, RNNs with continuously valued hidden layer representations were used for propagating the information over the sequence in studies such as~\cite{wan2017results},\cite{khorrami2017deep}, \cite{wu2017recent}.

\subsubsection{Tracking the Body Pose}

\begin{figure*}[h]
	\centering
	\includegraphics[width=\textwidth, height=3cm]{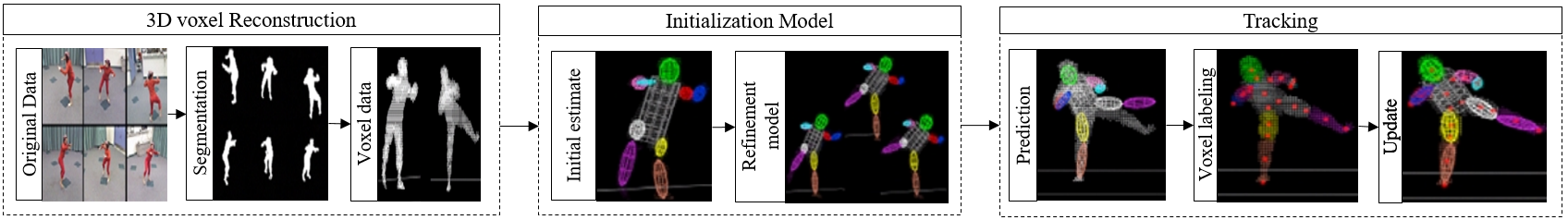}
	\caption{Example of dynamic body pose estimation. From left to right, we depict 3D voxel reconstruction segmentation with the segmentation and extracting the voxel data as input of the Initialitaion model and tracking box. Prediction by using the special filtering, voxel labeling and finally updating by means of the used filtering in the prediction step ~\cite{mikic2003human}.}
	\label{fig:dynampose}
\end{figure*}

Human actions can differ greatly in their dynamics. According to~\cite{wang2007learning}, they can be either periodic (e.g. running, walking or waving) or nonperiodic (e.g. bending), and either stationary (e.g. sitting) or nonstationary/transitional (e.g. skipping as a horizontal motion, and jumping or getting up as vertical motions). In videos we would like to track the human pose along a sequence of consecutive frames. In this way the parameters need to be estimated for every frame, i.e. the position, shape and configuration of the body are found such that they are consistent with the position and configuration of the body at the beginning of the motion capture process~\cite{mikic2003human}.

One way to facilitate the tracking of the human body pose is to require subjects to wear special markers, suits or gloves. Therefore, removing the constraints and the requirement of extra hardware were investigated by many researchers, in order to make it possible to perform tracking by using a single camera~\cite{mikic2003human}.

After finding the model for the first frame, for each of the next frames, it needs to be readjusted. A common approach to achieve the foregoing goal is to perform iterations such that every time, the model is predicted for the next frame. Expectation Maximization~(EM) can be utilized in order to create a human body tracking system, which assigns foreground pixels to the body parts, and then updates their positions according to the data~\cite{hunter1999visual,hunter1997estimation}. The parameters of the human body model can be projected onto the optical flow data based on the products of an exponential map~\cite{bregler1998tracking}. Optical flow and intensity can be used to find human body contours~\cite{delamarre20013d}. Then forces have to be applied in order to align the model with the contours extracted from the data. The foregoing process is iterated until the results converge. Configuration spaces of human motions with high dimensions can be handled using a particle filter model~\cite{deutscher2000articulated}. A continuation principle should be utilized based on annealing, in order to introduce the effect of narrow peaks into the fitness function. 

An intensely used method for modelling the temporal dimension of the human body is the use of probabilistic directed graphs like Hidden Markov Models (HMM). According to~\cite{bernhardt2010emotion}, HMMs were used in body gesture recognition~\cite{wilson1999parametric,wu1999vision}, automatic sign language interpretation~\cite{starner1997real} and activity analysis~\cite{mori2004hierarchical,oliver2002layered}. Despite the earlier works, which were based on static pattern matching and dynamic time warping~\cite{wu1999vision}, the most recent studies utilize HMMs and dynamic Bayesian networks~\cite{du2006recognizing,el2005generalization}.

Numerous methods of human body tracking may require manual initialization~\cite{mikic2003human}. Otherwise, usually it is necessary to perform a set of calibration movements, in order to  help the system to identify the body parts~\cite{kakadiaris1998three,cheung2000real}.
Other possibilities exist. 

% Dynamic pose estimation with LSTM ?  

% Future prediction, generative
%A trending recent topic is  \cite{oliu2017folded}

\subsubsection{3D Body Pose Detection and Tracking}

Instead of working on images, voxelized 3D models of human body shapes at each frame can be considered for the purpose of tracking~\cite{mikic2003human}. Voxels provide data, which can be used to create efficient and reliable tracking systems. However, their main disadvantage lies in demanding the additional pre-processing task of converting images to 3D voxel reconstructions, which requires dedicated hardware for a real-time performance~\cite{mikic2003human}.

In case of some systems, the input data were obtained using a single camera~\cite{sminchisescu2001covariance,ioffe2001human}. Other studies such as \cite{kakadiaris1996model,bregler1998tracking,delamarre20013d} proposed to use multiple cameras. A 3D kinematic model of a hand can be tracked based on a layered template representation of self-occlusions~\cite{rehg1995model}.

A formulated fully recursive framework for computer vision called DYNA was developed based on 2D blob features from multiple cameras, which were incorporated into a 3D human body model in a probabilistic way~\cite{wren2000understanding}. The system provided feedback from the 3D body model to the 2D feature tracking, where the prior probabilities were set using an extended Kalman filter. The foregoing approach can handle behaviors, which are meaningful motions, but can not be explained by passive physics. One example of this kind of tracking is represented in the Fig.~\ref{fig:dynampose}.

Tapered super-quadrics can be used to create human body models, and track one or more persons in 3D~\cite{gavrila19963}. A constant acceleration kinematic model should be used to predict the position of the body parts in the next frame. The undirected normalized chamfer distances between the image and model contours can be considered to adjust the positions of the head and torso and then those of the arms and legs.
%\textcolor{blue}{Other body pose detection methods exist, which have been less commonly utilized, e.g. the use of silhouette information \cite{senior2003real} or the usage of motion models for human body pose detection \cite{urtasun20043d}. The reason is that they possess relatively weak performances, are not highly flexible, and may require additional calculations and processing compared to the rest of the methods we have reviewed. Moreover, they might demand manual contribution from the user. For example, in~\cite{sabzmeydani2007detecting}, pedestrian detection from still images based on silhouette information was proposed. They used low-level gradient data to create, select and learn shapelet features. Although the efficiency of the system was relatively high, later, in studies such as~\cite{rogez2008randomized}, it was shown to result in a less accurate performance than approaches which utilize e.g. HOG features through random forest (RF)-based classification. On the other hand, as the main drawbacks of motion models, they are limited to a certain number of motions, allow only small variations in motions, and cannot handle transitions between different motions~\cite{moeslund2006survey}.} %For brevity we will not describe such techniques in detail. The interested reader should refer to surveys focused on this specific topic for further information.
Alternative less used pose detection methods are based on silhouette information \cite{senior2003real} or motion models \cite{urtasun20043d}. Their reported performances are weak and are not highly flexible, and usually may require additional calculations and processing compared to the rest of the methods we have reviewed. Moreover, they might demand manual contribution from the user. For example, in~\cite{sabzmeydani2007detecting}, pedestrian detection from still images based on silhouette information was proposed. They used low-level gradient data to create, select and learn shapelet features. Although the efficiency of the system was relatively high, later, in studies such as~\cite{rogez2008randomized}, it was shown to result in a less accurate performance than approaches which utilize e.g. HOG features through random forest (RF)-based classification. On the other hand, as the main drawbacks of motion models, they are limited to a certain number of motions, allow only small variations in motions, and cannot handle transitions between different motions~\cite{moeslund2006survey}. %For brevity we will not describe such techniques in detail. The interested reader should refer to surveys focused on this specific topic for further information.

\subsection{Representation Learning and Emotion Recognition}
\label{sec:representation_and_recognition}
The final stage of an EBGR process is building a relevant representation and using it to learn a mapping to the corresponding targets. Depending on the nature of the input, the representation can be static, dynamic or both. Also representation can be geometrical or could include appearance information and can focus on different parts of the body. Moreover, the mapping will then need to be taken into account in order to decide on the most probable class for a given input sample, i.e. to recognize it, which can be performed by using various classification methods. The foregoing topics will be discussed in what follows.

\subsubsection{Representation Learning}

Gunes et al.~\cite{gunes2004face,gunes2005affect} detected face and the hands based on the skin color information, and the hand displacement to neutral position was calculated according to the motion of the centroid coordinates. They used the information from the upper body. For example, in a neutral gesture, there is no movement, but in a happy or sad gesture, the body gets extended, and the hands go up, and get closer to the head than normal. More clearly, they defined motion protocols in order to distinguish between the emotions. 
In the first frame, the body was supposedly in its neutral state, i.e. the hands were held in front of the torso. In the subsequent frames, the in-line rotations of the face and the hands were analyzed. The actions (body modeling) were first coded by two experts. The first and the last frames from each body gesture, which stand for neutral and peak emotional states, respectively, were utilized for training and testing. 

Vu et al.~\cite{vu2011emotion} considered eight body action units, which represent the movements of the hands, head, legs and waistline. Kipp et al.~\cite{kipp2009gesture} provided an investigation of a possible correlation between emotions and gestures. The analysis was performed on static frames extracted from videos representing certain emotional states, as well as emotion dimensions of pleasure, arousal and dominance. The hand shape, palm orientation and motion direction were calculated for all the frames as the features. The magnitudes and directions of the correlations between the expected occurrences and the actual ones were evaluated by finding the correspondences between the dimension pairs, and calculating the resulting deviations. 

Glowinski et al.~\cite{glowinski2008technique} focused on the hands and head as the active elements of an emotional gesture. The features were extracted based on the attack and release parts of the motion cue, which refer to the slope of the line that connects the first value to the first relative extremum and the slope of the line that connects the last value to the last relative extremum, respectively. They also extracted the number of local maxima of the motion cue and the ratio between the maximum and the duration of the largest peak, which were used to estimate the overall impulsiveness of the movement. 

Kessous et al.~\cite{kessous2010multimodal} extracted the features from the body and hands. Based on silhouette and hands blobs, they extracted the quantity of motion, silhouette motion images (SMIs) and the  contraction index (CI).  Velocity, acceleration and fluidity of the hand’s barycenter were also computed. Glowinski et al.~\cite{glowinski2015towards} successfully extended their work using the same database as in~\cite{glowinski2008technique}, where the 3D position, velocity, acceleration and jerk were extracted from every joint of the skeletal structure of the arm. Kipp and Martin~\cite{kipp2009gesture} used a dimensional method to represent an affect emotional gesture along a number of continuous axes. Three independent bipolar dimensions namely, pleasure, arousal and dominance, were considered in order to define the affective states. The locations of 151 emotional terms were obtained.

In~\cite{castellano2008movement}, dynamic features were extracted in order to obtain a description of the submotion characteristics, including initial, final and main motion peaks. It was suggested that the timing of the motions greatly represents the properties of emotional expressions. According to~\cite{bernhardt2010emotion}, these features can be handled based on the concept of motion primitives, i.e. dynamic features can be represented by a number of subactions. 

Hirota et al.~\cite{hirota2011multimodal} used the information about the hands, where dynamic time warping (DTW) was utilized to match the time series. Altun et al.~\cite{altun2015recognizing} considered force sensing resistor (FSR) and accelerometer signals for affect recognition. Lim et al.~\cite{lim2014mei} captured 3D points corresponding to 20 joints at 30 frames per second (fps), where in the recognition stage, 100 previous frames were analyzed in case of every frame. 

Saha et al.~\cite{saha2014study} created skeleton models representing the 3D coordinates of 20 upper body joints, i.e. 11 joints corresponding to the hands, head, shoulders and spine were considered in order to calculate nine features based on the distances, accelerations and angles between them. The distance between the hands and spine, the maximum acceleration of the hands and elbows and the angle between the head, shoulder center and spine were considered as features, making use of static and dynamic information simultaneously. 

Camurri et al.~\cite{camurri2004toward} utilized five motion cues, namely, QoM, CI, velocity, acceleration and fluidity. Piana et al.~\cite{piana2013set} proposed 2D and 3D features for dictionary learning. The 3D data were obtained by tracking the subjects, and the 2D data from the segmentation of the images. The spacial data included 3D CI, QoM, motion history gradient (MHG) and barycentric motion index (BMI). Patwardhan et al.~\cite{patwardhan2016augmenting} utilized 3D static and dynamic geometrical features (skeletal) from the face and upper-body. Castellano et al.~\cite{castellano2007recognising} considered the velocity and acceleration of the trajectory followed by the hand's barycenter, which was extended in~\cite{castellano2008emotion}, adopting multiple modalities (face, body gesture, speech), and in~\cite{glowinski2008technique}, considering 3D features instead. Vu and et al.~\cite{vu2011emotion} used the AMSS~\cite{nakamura2008amss} in order to find the similarity between the gesture templates and the input samples.

Unfortunately more complex representations are very scarce in emotional body gesture recognition. Chen et al.~\cite{chen2013recognizing} used HOG on the motion history image (MHI) for finding the direction and speed, and Image-HOG features from bag of words (BOW) to compute appearance features. Another example is the usage of a multichannel CNN for learning a deep representation from the upper part of the body ~\cite{barros2015multimodal}. Finally, Botzheim et al.~\cite{botzheim2014gestural} used spiking neural networks for temporal coding. A pulse-coded neural network approximated the dynamics with the ignition phenomenon of a neuron and the propagation mechanism of the pulse between neurons. 

\subsubsection{Emotion Recognition}

Glowinski et al.~\cite{glowinski2015towards} showed that meaningful groups of emotions, related to the four quadrants of the valence/arousal space, can be distinguished from representations of trajectories of head and hands from frontal and lateral view of the body. A compact representation was grouped into clusters and used for classifying input into four classes according to the position in the dimensional space namely  high-positive(amusement, pride), high-negative (hot-anger, fear, dispair), low-negative (pleasure, relief, interest) or low-negative (cold anger, anxiety, sadness).

Gunes and Piccardi~\cite{gunes2005affect} used naive representations from the upper-body to classify body gestures into 6 emotional categories. The categories were groups of the original output space namely: anger-disgust, anger-fear, anger-happiness, fear-sadness-surprise, uncertainty-fear-surprise and uncertainty-surprise. The amount of data and subject diversity were low (156 samples, 3 subjects). A set of standard classifiers were trained and a Bayesian Net provided the best classification results.

Castellano et al.~\cite{castellano2007recognising} showed comparisons between different classifiers like 1-nearest-neighbor with dynamic time warping (DTW-1NN), J48 decision tree and the Hidden Naive Bayes (HNB) for classifying dynamic representations of body gestures as Anger, Joy, Pleasure or Sadness. The DTW-1NN provided the best results. 

Saha et al.~\cite{saha2014study} identified gestures corresponding to five basic human emotional states, namely, anger, fear, happiness, sadness and relaxation from skeletal geometrical features. They compared binary decision tree (BDT), ensemble tree (ET), k-nearest neighbour (KNN) and SVM, obtaining the best results by using ET.

Many studies modeled parts of the body independently for action analysis~\cite{bernhardt2010emotion}. For example, in~\cite{lv2006recognition}, actions based on arms, head and torso were modeled independently. On the contrary, a structural body model was proposed in~\cite{vacek2005classifying}. They defined a tree-based description of the body parts, where each activity corresponded to a node, based on the parts engaged in performing it. 

Context information such as background were introduced by Kosti et al.~\cite{kosti2017emotion}. They used a two low-rank filter CNN for extracting features from both body and background and fusing them for recognizing 26 emotions and intensity values of valence, arousal and dominance. Some examples of the emotions they targeted are peace, affection, fatigue and pain.

Although body gestures are important part of human communication, often they are a supplement of other reflexive behavior forms such as facial expression speech, or context. Studies in applied psychology showed that human recognition of facial expressions is influenced by the body expression and by the context \cite{van2007body}. Integration of verbal and nonverbal communication channels creates a system in which the message is easier to understand. Expanding the focus to several expression forms can facilitate research on emotion recognition as well as human-machine interaction.

% Speech and gestures
In literature, there are just a few examples concerning speech and gestures recognition. In \cite{yang2014analysis} the authors focused on uncovering the emotional effect on the interrelation between speech and body gestures. They used prosody and mel-frequency cepstral coefficient (MFCC) \cite{noroozi2016fusion,noroozi2017audio} for speech with three types of body gestures: head motion, lower and upper body motions to study the relationship between the two communication channels affected by the emotional state. Additionally, they proposed a framework for modeling the dynamics of speech-gesture interaction. 

In \cite{vu2011emotion} the authors also presented a bi-modal approach (gestures and speech) for recognition of four emotional states: happiness, sadness, disappointment, and neutral. Gestures recognition module fused two sources: video and 3D acceleration sensors. Speech recognition module was based on Julius \cite{webjulius} - open source software. The outputs from speech and gestures based recognition were fused by using weight criterion and best probability plus majority vote fusion methods. Fifty Japanese words (or phrases) and 8 types of gestures recorded from five participants were used to validate the system. Performance of the classifier indicated better results for bi-modal than each of the uni-modal recognition system. 

Regarding fusing gestures and faces the literature is also scarce. Gunes et al.~\cite{gunes2007bi} proposed a bi-model emotion recognition method from body and face. Features were extracted from two streams of video of the face and the body and fusions were performed at feature and decision level improving results with respect to individual feature space. A somehow similar approach was proposed by Caridakis et al. \cite{caridakis2007multimodal} combining face, body and speech modalities and early and late-fusion followed by simple statistical classification approaches with considerable improvement in results. 

Psaltis et al.~\cite{psaltis2016multimodal} introduced a multi-modal late fusion structure that could be used for stacked generalization on noisy databases. Surprise, hapiness, anger, sadness and fear were recognized from facial action units and high representation of the body gestures. The multi-modal approach provided better recognition than results from each of the mono-modal. 

A very interesting study on multi-modal automatic emotion recognition was presented in \cite{kessous2010multimodal}. The authors constructed a database consisting of audio-video recordings of people interacting with an agent in a specific scenario. Ten people of different gender, using several different native languages including French, German, Greek and Italian pronounced a sentence in 8 different emotional states. Facial expression, gesture and acoustic features were used with an automatic system based on a Bayesian classifier. Results obtained from each modality were compared with the fusion of all modalities. Combining features into multi-modal sets resulted in a large increase in the recognition rates, by more than 10\% when compared to the most successful unimodal system. Furthermore, the authors proved that the best results were obtained for gesture and speech feature merger.

%The authors of~\cite{ranganathan2016multimodal} presented the emoFBVP database, consisting of face, body gesture, voice and physiological signals. It contains recordings of actors displaying three different intensities of expressions of 23 different emotions. Using four deep belief network models generating robust multimodal features for emotion classification, they achieved better results than the state of the art methods. They proposed convolutional deep belief network models that learn salient features from multimodal feature sets of emotions expressions. This method gave better results when recognizing low intensity or subtle expressions of emotions compared to state of the art methods.
%Another approach was presented in~\cite{poria2015towards}, where the authors proposed a multimodal affective data analysis framework based on semantic and affective information (text, audio and video). In preliminary experiments using the eNTERFACE dataset, they achieved an accuracy of 87.95\%, outperforming the best state-of-the-art multimodal system by more than 10\%. 

\subsection{Applications}
\label{sec:applications}
Applications of automatic recognition of gesture based expression of affect are mainly of three types \cite{picard1997affective,picard1999affective,picard2010affective}. The first type consists of systems that detect the emotions of the users. The second type includes actual or virtual animated conversational agents, such as robots and avatars. They are expected to act similarly to humans when they are supposed to have a certain feeling. The third type includes systems that really feel the emotions. For example, these systems have applications in video telephony~\cite{cassell1998framework}, video conferencing and stress-monitoring
tool, violence detection ~\cite{pantic2000automatic,pantic2003toward,bermejo2011violence}, video surveillance~\cite{pentland2000looking}, and animation or synthesis of life-like agents~\cite{pantic2003toward} and automatic psychological research tools~\cite{pentland2000looking}. All the three types have been extensively discussed in the literature. However, this paper concentrates on affect detection only.

Automatic multi-modal emotion recognition systems can utilize sources of information that are based on face, voice and body gesture, at the same time. Thus they can constitute an important element of perceptual user interfaces, which may be utilized in order to improve the ease of use of online shops. They can also have applications in pervasive perceptual man-machine interfaces, which are used in intelligent affective machines and computers that understand and react to human emotions~\cite{picard2001toward}. If the system is capable of combining the emotional and social aspects of the situations for making a decision based on the available cues, it can be a useful assistant for humans~\cite{reeves1996people}.

%%%*********************** Data***********************************
\section{Data}
\label{sec:data}
\begin{figure}
    \label{fig:data}
	\centering
	\subfigure[]{
		\includegraphics[width=.45\linewidth, height=.35\linewidth]{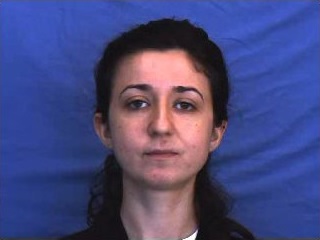}
		\label{fig:subfig1}
	}%
	\subfigure[]{
		\includegraphics[width=.45\linewidth, height=.35\linewidth]{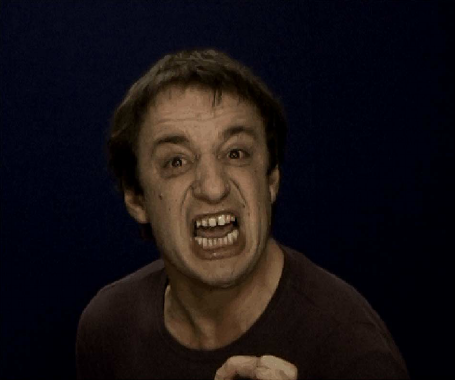}
		\label{fig:subfig2}
	}
	\subfigure[]{
		\includegraphics[width=.45\linewidth, height=.35\linewidth]{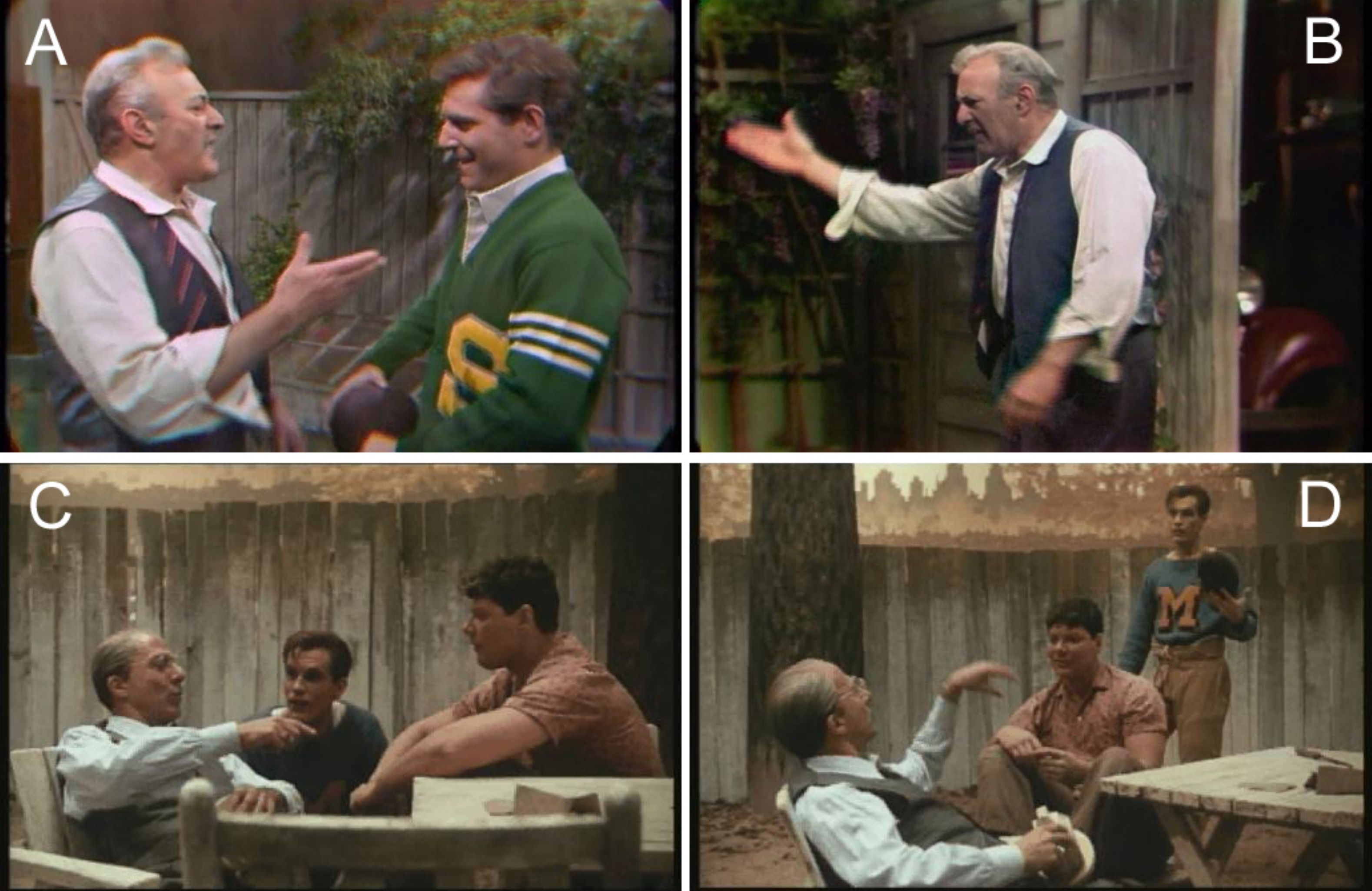}
		\label{fig:subfig3}
	}%
	\subfigure[]{
		\includegraphics[width=.45\linewidth, height=.35\linewidth]{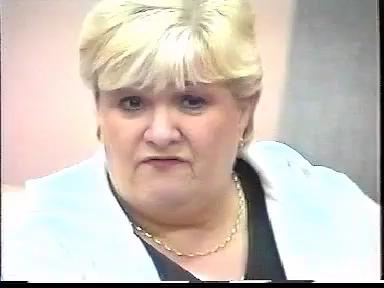}
		\label{fig:subfig4}
	}
	\subfigure[]{
		\includegraphics[width=.45\linewidth, height=.35\linewidth]{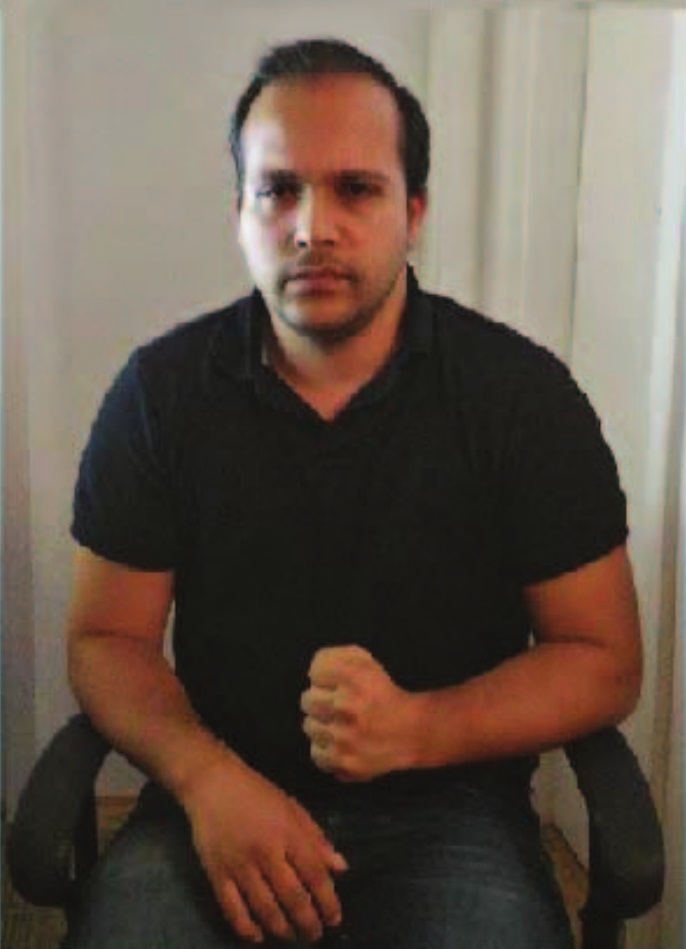}
		\label{fig:subfig5}
	}%
	\subfigure[]{
		\includegraphics[width=.45\linewidth, height=.35\linewidth]{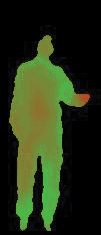}
		\label{fig:subfig8}
	}
	%\subfigure[]{
	%	\includegraphics[width=.45\linewidth, height=.35\linewidth]{MSRC_12}
	%	\label{fig:subfig6}
	%}
	%\subfigure[]{
	%	\includegraphics[width=.45\linewidth, height=.35\linewidth]{UCFKinect_1}
	%	\label{fig:subfig7}
	%}

	%\subfigure[]{
		%\includegraphics[height=.16\textheight]{emoFBVP}
		%\label{fig:subfig9}
	%}
	\caption{Selected samples from databases containing gesture based expressions of affect: (a)~FABO~\cite{gunes2006bimodal}, (b)~GEMEP-FERA~\cite{glowinski2008technique,baltruvsaitis2011real,valstar2012meta}, (c)~Theater~\cite{kipp2009gesture}, (d)~HUMAINE~\cite{castellano2007recognising,castellano2008emotion,kessous2010multimodal,douglas2011humaine}, 
	(e)~LIRIS-ACCEDE~\cite{gavrilescu2015recognizing,baveye2015liris}, 
	%(f) MSRC-12~\cite{fothergill2012instructing}, 
	%(g) UCFKinect~\cite{masood2011measuring},
	(f)~MSR-Action 3D~\cite{li2010action}.
	%(i) emoFBVP ~\cite{ranganathan2016multimodal}.
	}
\label{samples}
\end{figure}

We further present main public databases of gesture based expressions of affect useful for training EGBR systems. We discuss RGB, Depth and bi-modal of RGB + Depth databases in Sec. \ref{sec:data:rgb}, \ref{sec:data:depth} and \ref{sec:data:rgb-depth}, respectively. The reader is referred to Table \ref{tab:data} for an overview of the main characteristics of the databases and to Fig. \ref{samples} for a selection of database samples.
\subsection{RGB}
\label{sec:data:rgb}
One of the first body language databases with affect annotations was made publicly available by Gunes and Piccardi~\cite{gunes2005affect}. The database contains 206 samples with six basic emotions, as well as four more states, namely, neutral, anxiety, boredom and uncertainty. 156 samples were used for training, and 50 samples for the test.

Castellano et al.~\cite{castellano2007recognising,castellano2008emotion} collected affective body language data consisting of 240 gestures~\cite{humaine2008human}. Their database is a part of the HUMAINE database~\cite{douglas2011humaine}. There were six male and four female participants, i.e. 10 in total. They acted eight emotions, i.e. anger, despair, interest, pleasure, sadness, irritation, joy and pride, equally distributed in the valence arousal space. However, they focused on four emotions, i.e. anger, joy, pleasure and sadness. A camera filmed the full body of the subjects from the front view at a rate of 25 fps. In order to accelerate the silhouette extraction, they used a uniform dark background.

The Geneva Multi-modal Emotion Portrayals (GEMEP) database~\cite{glowinski2008technique} contains more than 7000 audio-video portrayals of emotional expressions. It includes 18 emotions portrayed by 10 actors. 150 portrayals were systematically chosen based on ratings by experts and non-experts, which resulted in the best recognition of the emotional intentions. Their analysis was on the basis of 40 portrayals selected from the mentioned set. They were chosen such that they represent four emotions, namely, anger, joy, relief and sadness. Each of these emotions is from one quadrant of the two main affective dimensions, i.e. arousal and valence.

The Theater corpus was introduced by Kipp and Martin~\cite{kipp2009gesture}, based on two movie versions of the play \textit{Death of a Salesman}, namely, DS-1 and DS-2.

Vu et al.~\cite{vu2011emotion} considered eight types of gestures that are present in the home party scenario of the mascot robot system. The database involved five participants, i.e. four males and one female. Their ages ranged from 22 to 30 years. They were from three different nationalities: Japanese, Chinese, and Vietnamese.

A subset of the LIRIS-ACCEDE video database~\cite{baveye2015liris} was created in~\cite{gavrilescu2015recognizing}, which contains upper bodies of 64 subjects, including 32 males and 32 females, with six basic emotions. Their ages were between 18 and 35 years. 

\subsection{Depth}
\label{sec:data:depth}
The GEMEP-FERA database, which was introduced by Baltru\v{s}aitis et al.~\cite{baltruvsaitis2011real}, is a subset of the GEMEP corpus. The training database was created by 10 actors. In the test database, six actors participated. From these actors, three were common with the training database, but the other three were new. The database consists of short videos of the upper body of the actors. The average length of the videos is 2.67 seconds. The videos do not start with a neutral state.

The database created by Saha et al.~\cite{saha2014study} involved 10 subjects. The age of the subjects ranged from 20 to 30. The subjects were stimulated by five different emotions, namely, anger, fear, happiness, sadness, and relaxation. These emotions caused the subjects to take different gestures accordingly. Each subject was filmed at a frame rate of 30 fps, for 60 seconds. Next, the Cartesian coordinates of the body joints were processed.

In~\cite{patwardhan2016augmenting}, six basic emotions, namely, anger, surprise, disgust, sad, happy and fear, were acted by 15 subjects. The subjects were between 25 to 45 years old. Five subjects were female, and the rest were male. In addition, five subjects were Americans, and the rest were Asians. The lighting conditions were controlled, and the poses of the bodies of the subjects were completely frontal. The subjects' distances from the camera were from to 1.5 to 4 meters.

The UCFKinect~\cite{masood2011measuring} was collected using Kinect and skeleton estimation from \cite{shotton2013real}. 16 subjects, including 13 males and three females, participated in the recordings. All the subjects were between 20 and 35 years old. Each of them performed 16 actions such as balance, punch or run and repeated it five times. In total, 1280 actions were recorded. The 3D coordinates of 15 joints were calculated for each frame. The data on the background and the clothes were not included in the calculations, and only the data on the skeleton was extracted.

The MSR Action 3D consists of twenty actions, namely, high arm wave, horizontal arm wave, hammer, hand catch, forward punch, high throw, draw x, draw tick, draw circle, hand clap, two hand wave, side-boxing, bend, forward kick, side kick, jogging, tennis swing, tennis serve, golf swing, pick up and throw.
\subsection{Bi-modal: RGB + Depth}
\label{sec:data:rgb-depth}
The database created by Psaltis et al.~\cite{psaltis2016multimodal} contains facial expressions that frequently appear in games. 
The subjects acted five basic emotions, namely, anger, fear, happiness, sadness and surprise. They considered a neutral emotion class for labeling the samples that do not present any motion, and cannot be classified under any of the basic emotions. 15 subjects participated to create 450 videos. Each video starts with an almost neutral state, and evolves toward a peak emotional state. The labels were assigned based on the emotion that the subject was asked to perform, not on the actually performed movements. The whole duration of every video is 3 seconds. Before acting each emotion, a video presenting the required emotion was shown to the subjects, and then the subjects performed the movements with their own styles five times. The database is divided into three parts. One part only contains facial expressions, another part only contains body gestures, and the last part contains both face and body data. They used a dense-ASM tracking algorithm for tracking the features and extracting the AUs. In order to evaluate the performance of the proposed method, they applied it to the FERA database as well.

The emoFBVP database~\cite{ranganathan2016multimodal} includes multi-modal recordings of actors, i.e. face, body gesture, voice and physiological signals. Audiovisual information of three different expressions, i.e. intensities, of 23 emotions are included, as well as tracking of facial features and skeletal tracking. 10 professional actors participated in acquiring the data. Each recording was repeated six times. Three recordings were in a standing position, and the others in a seated position. To date, this database offers the most diverse range of emotional body gestures in the literature, but right now it is not available.
Finally, the specifications of all available databases are summarized in Table~\ref{tab:data}. The list of emotions that have been considered in each of the databases is provided in Table~\ref{tab:labels}. A sample image from each of the databases can be seen in Fig. \ref{samples}.
\begin{table*}[t]
\centering
\caption{Main characteristics of a selected list of publicly available databases for recognizing gesture based expression of affect.}
\label{tab:data}
\resizebox{\textwidth}{!}{

\begin{tabular}{|c|c|c|c|c|c|c|c|c|c|c|c|c|c|c|}
\hline

\textbf{Reference} & \textbf{Name} & \textbf{Device} & \textbf{Body parts} & \textbf{Modality} & \rotatebox[origin=c]{-90}{\textbf{\#Emotions}} & \rotatebox[origin=c]{-90}{\textbf{\#Gestures}} & \rotatebox[origin=c]{-90}{\textbf{\#Subjects}} & \rotatebox[origin=c]{-90}{\textbf{\#Females}} & \rotatebox[origin=c]{-90}{\textbf{\#Males}} & \rotatebox[origin=c]{-90}{\textbf{\#Sequences}} & \rotatebox[origin=c]{-90}{\textbf{\#Samples}} & \rotatebox[origin=c]{-90}{\textbf{FR\footnote{Frame rate}} (fps)} & \textbf{Background} & \rotatebox[origin=c]{-90}{\textbf{AVL\footnote{Average video length}~(s)}}\\
\hline\hline

Gunes et al., 2006 \cite{gunes2006bimodal} & FABO & Digital camera & Face and body & Visual & 10 & NA & 23 & 12 & 11 & 23 & 206 & 15 & Uniform blue & $\sim$3600 \\ \hline

Glowinski et al., 2008 \cite{glowinski2008technique} & GEMEP & Digital camera & Face and body & Audiovisual & 18 & NA & 10 & 5 & 5 & 1260 & >7000 & 25 & Uniform dark & NA\\ \hline

%\multirow{2}{*}{Kipp et al., 2009~\cite{kipp2009gesture}} & Theater (DS-1) & \multirow{2}{*}{Camera} & \multirow{2}{*}{Whole body} & \multirow{2}{*}{Audiovisual} & \multirow{2}{*}{24} & 141 & \multirow{2}{*}{NA} & \multirow{2}{*}{NA} & \multirow{2}{*}{NA} & \multirow{2}{*}{NA} & 104 & \multirow{2}{*}{15} & \multirow{2}{*}{Nonuniform} & 1016\\
% & Theater (DS-2) & & & & & 117 & & & & & 69 & & & 528\\ \hline
 
Castellano et al., 2007 \cite{castellano2007recognising} & HUMAINE & Camera & Face and body & Audiovisual & 8 & 8 & 10 & 4 & 6 & 240 & 240 & 25 & Uniform dark & NA \\ \hline

Gavrilescu, 2015 \cite{gavrilescu2015recognizing} & LIRIS-ACCEDE & Camera & Face and upper body & Visual & 6 & 6 & 64 & 32 & 32 & NA & NA & NA & Nonuniform & 60 \\ \hline

Baltruvsaitis et al., 2011~\cite{baltruvsaitis2011real} & GEMEP-FERA & Kinect & Upper body & Visual & 5 & 7 & 10 & NA & NA & 289 & NA & 30 & Uniform dark & 2.67 \\ \hline

Fothergill et al., 2012~\cite{fothergill2012instructing} & MSRC-12 & Kinect & Whole body & Depth & NA & 12 & 30 & 40\% & 60\% & 594 & 6244 & 30 & Uniform white & 40 \\ \hline

Masood et al., 2011~\cite{masood2011measuring} & UCFKinect & Kinect & Whole body & Depth & NA & 16 & 16 & NA & NA & NA & 1280 & 30 & NA & NA\\ \hline

Li et al., 2010 \cite{li2010action} & MSR-Action 3D & Structured light & Whole body & Depth & NA & 20 & 7 & NA & NA & NA & 567 & 15 & Nonuniform & NA\\ \hline

%Ranganathan et al., 2010~\cite{ranganathan2016multimodal} & emoFBVP & Kinect & Face and body & Audiovisual & 23 & 23 & 10 & NA & NA & 1380 & 1380 & 30 & Nonuniform & NA\\ \hline
\end{tabular}
}
\end{table*}
% \usepackage{booktabs}
\begin{table}[]
\centering
\tiny
\caption{Labels included in a selected list of the databases. F = FABO \cite{gunes2006bimodal}, G = GEMEP \cite{glowinski2008technique}, T = T heater \cite{kipp2009gesture}, H = HUMAINE \cite{castellano2007recognising}, LA = LIRIS-ACCEDE \cite{gavrilescu2015recognizing}, GF = GEMEP-FERA \cite{baltruvsaitis2011real}.}
\label{tab:labels}
\resizebox{0.48\textwidth}{!}{%
\begin{tabular}{@{}|c|ccccccc|@{}}
\hline
\textbf{Database} & \textbf{F} & \textbf{G} & \textbf{T} & \textbf{H} & \textbf{LA} & \textbf{GF} & \textbf{Frequency} \\ \hline\hline
Sadness          & $\bullet$          & $\bullet$          & $\bullet$          & $\bullet$          & $\bullet$           & $\bullet$           & 6                  \\\hline
Anger            & $\bullet$          &            & $\bullet$          & $\bullet$          & $\bullet$           & $\bullet$           & 5                  \\\hline
Anxiety          & $\bullet$          & $\bullet$          & $\bullet$          &            &             &             & 3                  \\\hline
Disgust          & $\bullet$          & $\bullet$          &            &            & $\bullet$           &             & 3                  \\\hline
Fear             & $\bullet$          &            &            &            & $\bullet$           & $\bullet$           & 3                  \\\hline
Surprise         & $\bullet$          & $\bullet$          &            &            &             &             & 3                  \\\hline
Boredom          & $\bullet$          &            & $\bullet$          &            &             &             & 2                  \\\hline
Happiness        & $\bullet$          &            &            &            &             & $\bullet$           & 2                  \\\hline
Interest         &            & $\bullet$          &            & $\bullet$          &             &             & 2                  \\\hline
Contempt         &            & $\bullet$          &            &            &             &             & 2                  \\\hline
Despair          &            & $\bullet$          &            & $\bullet$          &             &             & 2                  \\\hline
Irritation       &            & $\bullet$          &            & $\bullet$          &             &             & 2                  \\\hline
Joy              &            &            &            & $\bullet$          &             & $\bullet$           & 2                  \\\hline
Pleasure         &            & $\bullet$          &            & $\bullet$          &             &             & 2                  \\\hline
Relief           &            & $\bullet$          &            &            &             & $\bullet$           & 2                  \\\hline
Admiration       &            & $\bullet$          & $\bullet$          &            &             &             & 1                  \\\hline
Neutral          &            & $\bullet$          &            &            &             &             & 1                  \\\hline
Pride            &            & $\bullet$          &            & $\bullet$          &             &             & 1                  \\\hline
Shame            &            & $\bullet$          &            &            &             &             & 1                  \\\hline
Aghastness       &            & $\bullet$          &            &            &             &             & 1                  \\\hline
Amazement        &            &            & $\bullet$          &            &             &             & 1                  \\\hline
Amusement        &            & $\bullet$          &            &            &             &             & 1                  \\\hline
Boldness         &            &            &            & $\bullet$          &             &             & 1                  \\\hline
Comfort          &            &            &            & $\bullet$          &             &             & 1                  \\\hline
Dependency       &            &            & $\bullet$          &            &             &             & 1                  \\\hline
Disdain          &            &            & $\bullet$          &            &             &             & 1                  \\\hline
Distress         &            &            & $\bullet$          &            &             &             & 1                  \\\hline
Docility         &            &            & $\bullet$          &            &             &             & 1                  \\\hline
Elation          &            & $\bullet$          &            &            &             &             & 1                  \\\hline
Excitement       &            &            & $\bullet$          &            &             &             & 1                  \\\hline
Exuberance       &            &            & $\bullet$          &            &             &             & 1                  \\\hline
Fatigue          & $\bullet$          &            &            &            &             &             & 1                  \\\hline
Gratefulness     &            &            & $\bullet$          &            &             &             & 1                  \\\hline
Hostility        &            &            & $\bullet$          &            &             &             & 1                  \\\hline
Indifference     &            &            & $\bullet$          &            &             &             & 1                  \\\hline
Insecurity       &            &            & $\bullet$          &            &             &             & 1                  \\\hline
Nastiness        &            &            & $\bullet$          &            &             &             & 1                  \\\hline
Panic Fear       &            & $\bullet$          &            &            &             &             & 1                  \\\hline
Rage             &            & $\bullet$          &            &            &             &             & 1                  \\\hline
Relaxation       &            &            & $\bullet$          &            &             &             & 1                  \\\hline
Respectfulness   &            &            & $\bullet$          &            &             &             & 1                  \\\hline
Satisfaction     &            &            & $\bullet$          &            &             &             & 1                  \\\hline
Tenderness       &            & $\bullet$          &            &            &             &             & 1                  \\\hline
Uncertainty      & $\bullet$          &            &            &            &             &             & 1                  \\ \hline
Unconcern        &            &            & $\bullet$          &            &             &             & 1                  \\ 
\hline
\end{tabular}
}
\end{table}
\section{Discussion}
\label{sec:discussion}
In this section we discuss the different aspects of automatic emotional body gesture recognition presented in this work. We start with the collections of data currently available for the community. Then, the discussion mainly focuses on representation building and emotion recognition from gestures. This includes the categories of mostly used features, taking advantage of complementarity by using multi-modal approaches and most common pattern recognition methods and target spaces. 

\subsection{Data}
\label{sec:discusssion:data}
%Dorota
Majority of freely accessible data sets contain acted expressions. This type of material is usually composed of high quality recordings, with clear undistorted emotion expression. The easiness of acquiring such recordings opens a possibility of obtaining several samples from a single person. The conventional approach to collect acted body language databases is to let actors present a scene portraying particular emotional states. Professionals are able to immerse in an emotion they perform, which may be difficult for ordinary people. This kind of samples are free from uncontrollable influences and usually they do not require additional evaluation and labeling processes.
However, some researchers emphasize that these types of recordings may lead to creating a set of many redundant samples, as there is a high dependency on actors skills and his or her ability to act out the same emotional state differently. Another argument against such recordings states that they do not reflect real world conditions. Moreover, acted emotions usually comprise of basic emotions only, whereas in real life emotions are often weak, blurred, occur as combinations, mixtures, or compounds of primary emotions. Wherefore current trends indicate that spontaneous emotions are preferable for research. There is another method to record emotional body movements in natural situations. One can use movies, TV programs such as talk shows, reality shows or live coverage. This type of material might not always be of satisfactory quality (background noise, artifacts, overlapping, etc.) and may obscure the exact nature of recorded emotions. Moreover, collections of spontaneous samples must be evaluated by human decision makers or professional behaviorists to determine the gathered emotional states. Nonetheless it does not guarantee objective, genuinely independent assessments. Additionally, copyright reasons might make it difficult to use or disclose movies or TV recordings. An accurate solution for sample acquisition may be provoking an emotional reaction using staged situations, which has been already used in emotion recognition from speech or mimics. Appropriate states may be induced using imaging methods (videos, images), stories or computer games. This type of recordings are preferred by psychologists, although the method can not provide desirable effects as reaction to the same stimuli may differ. Similarly to spontaneous speech recordings, triggered emotional samples should be subjected to a process of labeling. Ethical or legal reasons often prohibit to use or make them publicly available. Taking into account above mentioned issues, real-life emotion databases are rarely available to the public, and a good way of creating and labelling such samples is still open to question.

The process of choosing appropriate representation of emotional states is intricate. It is still debatable how detailed and which states should be covered. Analyzing Table \ref{tab:labels} one can observe how  broad affective spectrum has been used in various types of research. Most authors focus on sets containing six basic emotions (according to Ekman’s model). Sadness and anger occur in majority of databases. Fear, surprise and disgust are also commonly used. However, there are quite a lot of affective states that are not consistently represented in the available databases. Some examples (see Tab. \ref{tab:labels}) are uncertainty, unconcern, aghastenss, shame, tenderness, etc. 

There is a lack of consistency in the taxonomy used for naming the affective states. For example both joy and happiness, are used interchangeably depending on the database. It is difficult to evaluate whether these are the same or different states. Joy is more beneficial, as it is less transitory than happiness and is not tied to external circumstances. Therefore, it is possible that there is a misunderstanding in naming: while happiness may be caused by down to earth experiences, material objects, joy needs rather spiritual experiences, gratitude, and thankfulness, thus may be difficult to evoke and act. These misunderstandings may be also a result of translations. Such issues will reoccur until a consistent taxonomy of emotions will be presented, so far there is no agreement among experts even on the very definition of primary states. Moreover, due to the heterogeneity of described databases, comparison of their quality is problematic. With just several public accessible emotional databases and with the addition of the above described issues, comparison of detection algorithms becomes a challenging task. There is clearly space and necessity of creation of more unified emotional state databases.

\subsection{Representation Learning and Emotion Recognition}

\iffalse
The performances of a few methods which use the whole or upper body for emotional gesture recognition are listed in Table~\ref{TD1}. It can be seen that using the whole body has always resulted in better performance. Some studies that have used various modalities are also listed in Table~\ref{T3}.
\begin{table}[t]
	\centering
	\caption{Comparison of the effect of using the whole body with the upper body for emotional gesture recognition.}
	\label{TD1}
	\resizebox{0.5\textwidth}{!}{%
		\begin{tabular}{@{}|c|c|c|c|c|@{}}
			\hline
			\textbf{Body parts}        & \textbf{\#emotions}  & \textbf{Performance~(\%)}  &\textbf{Reference} \\ \hline \hline                  Whole          & 8 & 92.70            &\cite{hirota2011multi-modal}      \\\hline
			Whole          & 5                   & 93.00            &\cite{psaltis2016multimodal}     \\\hline
			Whole          & 8                   & 94.06            &\cite{ellis2013exploring}        \\\hline
			Upper         & 6                   & 75.00            &\cite{gavrilescu2015recognizing} \\\hline
			Upper         & 5                   & 90.83            & \cite{saha2014study}            \\\hline
			Upper         & 6                   & 90.00            & \cite{gunes2005affect}          \\ \hline
		\end{tabular}
	}
\end{table}
\fi

 \begin{table}[t]
 \centering
 \caption{Summary of a few multi-modal emotion recognition methods. S=Speech, F=Face, H=Hands, B=Body. }
 \label{T3}
 \resizebox{0.5\textwidth}{!}{%
 \begin{tabular}{|l|c|c|c|c|}
 \hline 
 \multicolumn{1}{|c|}{\textbf{Reference}} & \textbf{Modalities} & \textbf{\#samples} & \textbf{\#emotions} & \textbf{Representation} \\ \hline \hline
 Gunes Piccardi~\cite{gunes2005affect} & F + B & 206 & 6 & Motion protocols \\ \hline
 Castellano's et al.~\cite{castellano2007recognising} & B & 240  & 4 & Multi cue \\\hline
 Castellano et al.~\cite{castellano2008emotion} & B  & 240  & 4 & Multi cues \\ \hline
 Glowinski et al.~\cite{glowinski2008technique} & B  & 40 & 4 & Multi cues \\\hline
 Kipp Martin~\cite{kipp2009gesture} & B & \#119 & 6 & PAD   \\ \hline
 Kessous et al.~\cite{kessous2010multimodal}          & S + F + B     & NA    & 8                 & Multi cues \\\hline
 Vu et al.~\cite{vu2011emotion} & S + B & 5 & 4 & Motion protocols \\ \hline
 Gavrilescu~\cite{gavrilescu2015recognizing} & B + H & 384 & 6 & Motion protocols    \\ \hline
 \end{tabular}}
 \end{table}

\textbf{Representation Learning}. The large majority of the methods developed to recognize emotion from body gestures use geometrical representations. A great part of these methods build simple static or dynamic features related to the coordinates of either joints of kinematic models or of parts of the body like head, hands or torso. Some of the most used features are displacements \cite{gunes2005affect}, orientation of hands \cite{kipp2009gesture} motion cues like velocity and acceleration  \cite{glowinski2008technique, glowinski2015towards, castellano2008movement, altun2015recognizing, camurri2004toward, castellano2007recognising}, shape information and silhoutte \cite{kessous2010multimodal}, smoothness and fluidity, periodicity, spatial extent and kinetic energy, among others. While most descriptors are very simple there are also examples of slightly more advanced descriptors like Quantity of Motion (QoM measures of the amount of motion in a sequence), Silhoutte Motion Images (SMI contains information about the changes of the shape and position of the silhouette), Contraction Index (CI measures the level of contraction or expansion of the body), and Angular Metrics for Shape Similarity (AMSS)  \cite{piana2013set, vu2011emotion}. 

% Dynamic + Static
Considering dynamic features such as acceleration, movement gain and velocity, or at least combining them with static features, usually leads to higher recognition rates than relying solely on the latter, since they result in a richer representation and since some emotional traits are expressed mostly in the dynamics of the human body.

\begin{figure}
\centering
\includegraphics[width=.40\textwidth, height=3cm]{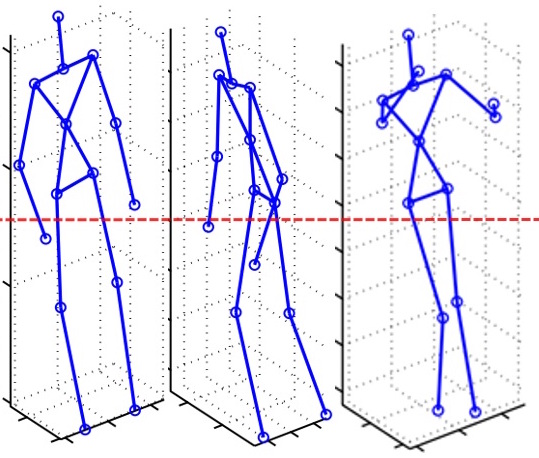}
\caption{Sample upper and lower body postures~\cite{ellis2013exploring}.}
\label{FD1}
\end{figure}

% Body parts
Most of the methods proposed focus on the upper body (head, neck, shoulders, arms, hands and hand fingers) \cite{kipp2009gesture, hirota2011multimodal, saha2014study},  hands \cite{glowinski2008technique}, arm \cite{glowinski2015towards}, body and hands \cite{kessous2010multimodal}, full body \cite{vu2011emotion,piana2013set}. Upper body and lower body parts are represented in Fig. \ref{FD1}. 

% Hands
Among all different parts of the body, in the context of body gesture recognition, numerous studies have focused on hand gestures, which requires hand segmentation and tracking. The features that can be extracted from the hands include palm orientation, hand shape, elbow, wrist, palms and shoulder joints, hand shape and motion direction, which are analyzed independently from the body, in order to calculate the motion of the hand and the individual fingers along each of the axes. The motions of the hand are measured with the body as the reference. The motions of the body itself are found in terms of the changes of the pose of the upper body, i.e. its inclinations to the left, right, forward or backward.

% Body parts
%Therefore, most papers focus only on the data from the hands, arms and head movements. However, while performing emotional gesture or posture recognition, some other body parts may unconsciously take certain forms that are useful to detect the emotional state more accurately. That is why in some studies, the whole body is considered for emotional gesture recognition, in order to take advantage of the information from the legs as well. The data from the legs might affect the resulting emotion label, i.e. the output of the classifier. Therefore, it might be necessary to create a database that contains data from the whole body, rather than using a database that already exists.

% DEEP 
Complex learnt representations for recognizing emotion from body gestures are very scarce, mostly because there is a lack of big volumes of labelled data (see Tab. \ref{tab:data}) for learning such representations in a supervised way. Two of the very few works that uses deep learning representations for body emotion recognition are multichannel CNN from upper body~\cite{barros2015multimodal} and spiking neural networks for temporal coding ~\cite{botzheim2014gestural}. As previously discussed in Sec. \ref{sec:discusssion:data} there is a lack of consistent taxonomy for the output spaces in the various databases published to date. This results in considerable fragmentation of the data and makes transfer learning techniques difficult. Even though not explored yet in the literature, unsupervised learning might be interesting for pre-training general representations of the moving human body, before tuning to more specific emotion oriented models.

\textbf{Emotion Recognition}. 
% Simplified ouput spaces
There is a tendency in the literature to reduce the output spaces for simplifying the recognition problem. This has been done either by grouping emotions into quadrants of a dimensional emotion space \cite{glowinski2015towards} or by grouping emotions based on similarity of their apperance \cite{gunes2005affect}. In general, most methods have focused on recognizing basic emotions like Anger, Joy, Pleasure, Sadness and Fear \cite{castellano2007recognising, saha2014study}. Though not dominant, methods that target richer output spaces also exist \cite{kosti2017emotion}. 

% Comparision btw different standard clasiffiers 
Another popular approach is to show extensive comparison between sets of standard classifiers like decision trees, k-NNs and SVMs.  The results of using numerous classifiers and different numbers of emotion classes based on two different databases are summarized in Table~\ref{TD2:1} and \ref{TD2:2}, respectively.

\begin{figure}[t]
\centering
\includegraphics[width=0.48\textwidth]{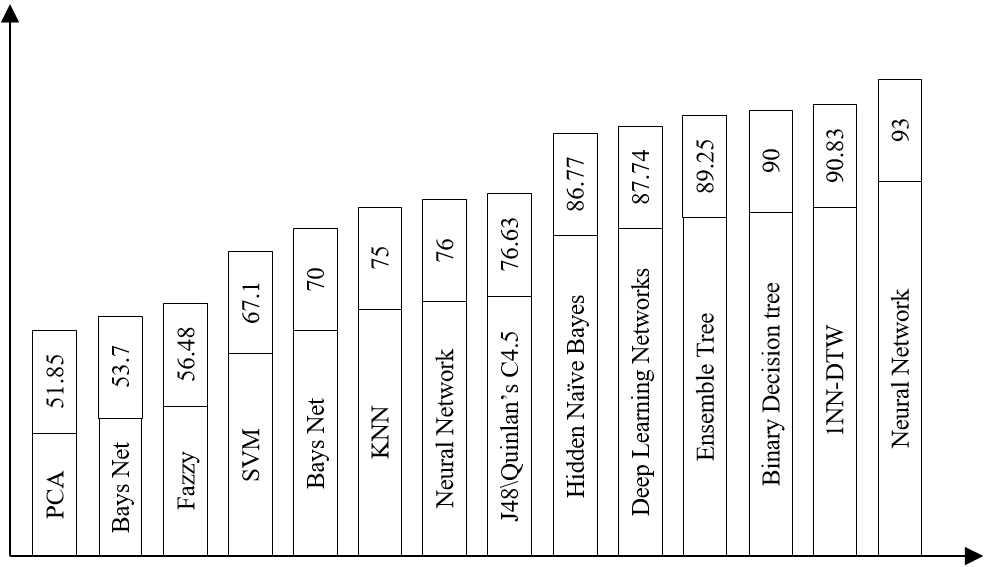}
\caption{Performances~(\%) of different emotion recognition methods based on the different databases.}
\label{FD2}
\end{figure}

According to Table~\ref{TD2:1}, J48 \cite{chauhan2013implementation} has the best performance between three used classifiers tested on HUMAINE database. This work used just four labels of the mentioned database. 
According to Table~\ref{TD2:2} the best performance on a different database, with 10 subjects and 5 labels, is achieved by ensemble tree classification methods. Moreover, the different methods with their performances are  represented as a chart for emotional gesture recognition in Fig. \ref{FD2}.

% Refine emotion recognition by using strctured learning approaches
More complete representations of the body can also be used in a more meaningful way. Particularly interesting are structural models where different parts of the body are independetly representated and contribute to a final decision over the emotion which takes into account predefined priors \cite{vacek2005classifying}. Going even further, additional information from the context, like the background could be used as well to refine final decision \cite{kosti2017emotion}.    

% Multimodality
Different body language components (gestures, faces) together with speech carry affective information and complementary processing have obvious advantages. A consistent part of the literature uses multiple representations of the body in a complementary way to recognize emotion. For example, there are works that combine body with speech \cite{yang2014analysis, vu2011emotion} and with face\cite{gunes2007bi, caridakis2007multimodal}. Regardless of the fusion techniques used, all these methods report improvements of results backing the hypothesis that there is considerable complementarity in different modalities and its exploration is fruitfull. Also, it has already been previously commented that a more complete body representation is also helpful in this respect (for example, upper and lower body considered together). Unfortunately research in multi-modal emotion recognition remains rather scarce and simplistic. The few works that exists mostly focus in simplistic fusion techniques from shallow representations of body and face or body and speech. Even though all methods report important improvements over monomodal equivalents, this potential remains largely unexplored. The reader is refered to Table~\ref{T3} for a selected set of studies that have used body representations together with representations of other modalities for recognizing emotion.

\begin{table}[t]
 \centering
 \caption{Comparison of the effect of using various classification methods and different numbers of emotion classes based on HUMAINE EU-IST database \cite{castellano2008emotion}.}
 \label{TD2:1}
\resizebox{0.48\textwidth}{!}{%
 \begin{tabular}{@{}|l|c|c|c|c|l|@{}}
 \hline
\textbf{Classifier}    & \textbf{Performance~(\%)} & \textbf{\#classes}       \\ \hline \hline
1NN-DTW                & 53.70              & 4                               \\\hline
J48 or Quinlan's C4.5  & 56.48              & 4                               \\\hline
Hidden Naive Bayes     & 51.85              & 4                              \\\hline
 \end{tabular}
}
\end{table}   

\begin{table}[t]
 \centering
 \caption{Comparison of the effect of using various classification methods and different numbers of emotion classes based on the recorded samples by Kinect. The database included by 10 subjects in the age group of 25$\pm $5 years \cite{saha2014study}.}
 \label{TD2:2}
\resizebox{0.48\textwidth}{!}{%
 \begin{tabular}{@{}|l|c|c|c|c|c|l|@{}}
 \hline
\textbf{Classifier}    & \textbf{Performance~(\%)} & \textbf{\#classes}             \\ \hline \hline
Ensemble tree          & 90.83              & 5                                      \\\hline
Binary decision tree,  & 76.63              & 5                                      \\\hline
K-NN                   & 86.77              & 5                                       \\\hline
SVM                    & 87.74              & 5                                       \\\hline
Neural network         & 89.26              & 5                                       \\\hline 
 \end{tabular}
}
\end{table}   

The number of emotion classes affects the performance of a given classifier as well. Usually, reducing the number of classes from a given database should increase the performance. %According to Fig.~\ref{FD3}, 
The best recognition rate, i.e. 93\%, is obtained by considering five emotion classes and using neural networks. It should be noted that low-quality samples or features may degrade the performance, and cause a violation of the expected trend.

%\begin{figure}[t]
%\centering
%\includegraphics[width=0.48\textwidth]{class_performance.jpg}
%\caption{The performance rate~(\%) as a %function of the number of emotion classes.}
%\label{FD3}
%\end{figure}
%%%%% Because some one has been disable the figures, ->
%The performances of different classifiers have been drawn in Fig.~\ref{FD4}, where the numbers of emotion classes have been shown as well.

\iffalse
\begin{figure}[t]
\centering
\includegraphics[width=0.48\textwidth]{totall3_2.jpg}
\caption{The performance rates~(\%) of different classifiers, as well as the numbers of emotion labels.}
\label{FD4}
\end{figure}
\fi

Approaches to train and test emotional gesture recognition systems are investigated based on the existing literature, where a certain portion of the database is used for training, and the rest is left for testing. Some of the proposed techniques present superior performances on specific databases, i.e. they have led to accuracy rates higher than $90\%$. However, in order to ensure that the system is reliable, it needs to be tested against different types of data, including various conditions of background, e.g. dark, light, uniform and nonuniform. Moreover, it is worth paying attention that different training and testing strategies may result in different performance rates.

%%%*********************** Conclusion***********************************
\section{Conclusion}
\label{sec:conclusion}
In this paper we defined a general pipeline of Emotion Body Gesture Recognition methods and detailed its main blocks. We have briefly introduced important pre-processing concepts like person detection and body pose estimation and detailed a large variety of methods that recognize emotion from body gestures grouped along important concepts such as representations learning and emotion recognition methods.  For introducing the topic and broadening its scope and implications we defined emotional body gestures as a component of body language, an essential type of human social behavior. The difficulty and challenges of detecting general patterns of affective body language are underlined. Body language varies with gender and has important cultural dependence vital issues for any researcher willing to publish data or methods in this field.   

In general the representations used remain shallow. Most of them are naive geometrical representations, either skeletal or based on independently detected parts of the body. Features like motion cues, distances, orientations or shape descriptors abound. Even though recently we can see deep meaningful representations being learned for facial analysis for affect recognition a similar approach for a more broader affective expression of humans is still to be developed in the case of body analysis. For sure the scarcity of body gesture and multimedia affective data is playing a very important role, problem that recently is starting to be overcome in the case of facial analysis. An additional problem is that while in the case of facial affective computing there has been a quite clear consensus of the output space (primitive facial expressions, facial Action Units and recently more comprehensive output spaces) in the case of general affective expressions in a broader sense such consensus does not exist. A proof in this sense is the variety of labels proposed in the multitude of publicly available data, some of them following redundant or confusing taxonomies. 

In general, for comprehensive affective human analysis from body language, emotional body gesture recognition should learn from emotional facial recognition and clearly agree on sufficiently simple and well defined output spaces based on which to publish large high quality amounts of labelled and unlabelled data that could serve for learning rich deep statistical representations of the way the affective body language looks like.   

\ifCLASSOPTIONcompsoc
  \section*{Acknowledgments}
\else
  \section*{Acknowledgment}
\fi

This work is supported Estonian Research Council Grant (PUT638), Estonian-Polish Joint Research Project, the Estonian Centre of Excellence in IT (EXCITE) funded by the European Regional Development Fund, the Spanish Project TIN2016-74946-P (MINECO/FEDER, UE), CERCA Programme / Generalitat de Catalunya and the Scientific and Technological Research Council of Turkey (T\"UBİTAK) (Proje 1001 - 116E097). %This project has received funding from the European Union’s Horizon 2020 research and innovation programme under the Marie Sklodowska-Curie grant agreement Nº 665919. We gratefully acknowledge the support of NVIDIA Corporation with the donation of the Titan Xp GPU used for this research.

\ifCLASSOPTIONcaptionsoff
  \newpage
\fi
 \bibliography{ms}
 \bibliographystyle{elsarticle-num}

\begin{IEEEbiography}[{\includegraphics[width=1in,height=1.25in,clip,keepaspectratio]{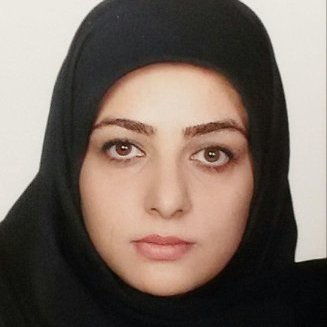}}] {Fatemeh Noroozi} received her B.Sc. in Computer Engineering, Software, from Shiraz University, Iran. Her thesis was entitled \textquotedblleft Modeling of Virtual Organizations Integrating on Physical Core based on a
Service-oriented Architecture\textquotedblright. Afterwards, she received her M.Sc. in Mechatronics Engineering from the University of Tehran, Iran. Her thesis was entitled \textquotedblleft Developing a Real-time Virtual Environment for Connecting to a Touching Interface in Dental Applications \textquotedblright. Currently, she is a PhD student at the University of Tartu, Estonia, working on \textquotedblleft multi-modal Emotion Recognition based Human-robot Interaction Enhancement\textquotedblright.

\end{IEEEbiography}

 \begin{IEEEbiography}[{\includegraphics[width=1in,height=1.25in,clip,keepaspectratio]{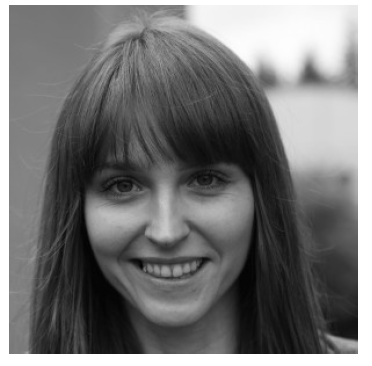}}] {Dorota Kamińska} graduated in Automatic Control and Robotics and completed postgraduate studies in Biomedical
image processing and analysis at Łodz University of
Technology. She received her PhD degree from Faculty
of Electrical, Electronic, Computer and Control Engineering
at Łodz University of Technology in 2014. The topic
of her thesis was “Emotion recognition from spontaneous
speech”. She gained experience during the TOP 500 Innovators
programme at Haas School of Business, University
of California in Berkeley. Currently she is an educator
and scientist at Institute of Mechatronics and Information
Systems. She is passionate about biomedical signals
processing for practical appliances. As a participant
of many interdisciplinary and international projects, she is
constantly looking for new challenges and possibilities of
self-development.
\end{IEEEbiography}

 \begin{IEEEbiography}[{\includegraphics[width=1in,height=1.25in,clip,keepaspectratio]{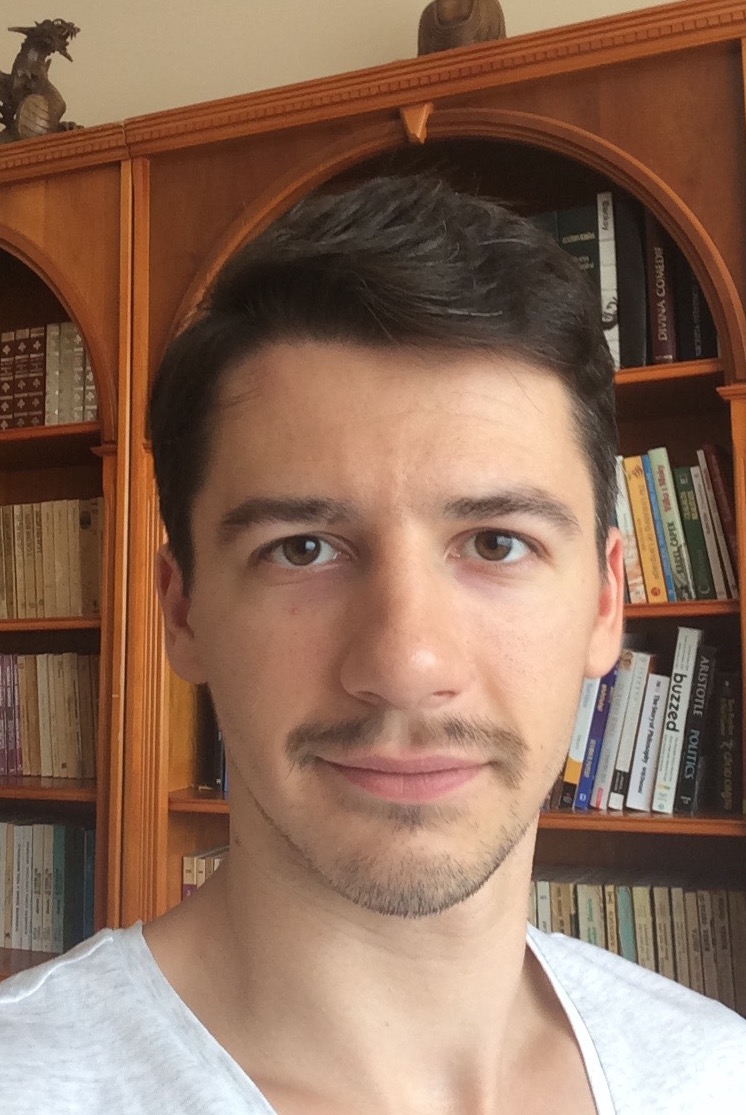}}]{Ciprian Adrian Corneanu} got his BSc in Telecommunication Engineering from T\'{e}l\'{e}com SudParis in 2011 and his MSc in Computer Vision from Universitat Aut\'{o}noma de Barcelona in 2015. Currently he is a Ph.D. student at the Universitat de Barcelona and a fellow of the Computer Vision Center, UAB. His main research interests include face and behavior analysis, affective computing, social signal processing and human computer interaction.
\end{IEEEbiography}

\begin{IEEEbiography}[{\includegraphics[width=1in,height=1.25in,clip,keepaspectratio]{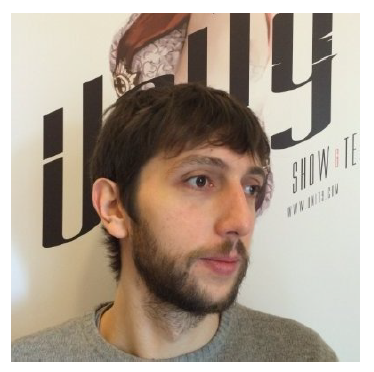}}] {Tomasz Sapiński} received his M.Sc. degree in Computer Science from Faculty of Technical Physics, Information
Technology and Applied Mathematics at Łodz University
of Technology. Currently he is Ph.D. student at Institute of
Mechatronics and Information Systems, Łodz University
of Technology. His main research topics are: multi-modal
emotion recognition and practical applications of virtual
reality.
\end{IEEEbiography}

\begin{IEEEbiography}[{\includegraphics[width=1in,height=1.25in,clip,keepaspectratio]{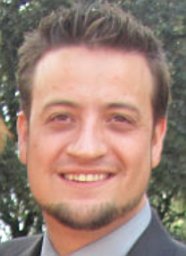}}]{Sergio Escalera}
obtained the P.h.D. degree on Multi-class visual categorization systems at Computer Vision Center, UAB. He obtained the 2008 best Thesis award on Computer Science at Universitat Autònoma de Barcelona. He leads the Human Pose Recovery and Behavior Analysis Group at UB, CVC, and the Barcelona Graduate School of Mathematics. He is an associate professor at the Department of Mathematics and Informatics, Universitat de Barcelona. He is an adjunct professor at Universitat Oberta de Catalunya, Aalborg University, and Dalhousie University. He has been visiting professor at TU Delft and Aalborg Universities. 
He is also a member of the Computer Vision Center at UAB. He is series editor of The Springer Series on Challenges in Machine Learning. 
He is vice-president of ChaLearn Challenges in Machine Learning, leading ChaLearn Looking at People events. His research interests include, between others, statistical pattern recognition, affective computing, and human pose recovery and behavior understanding, including multi-modal data analysis.

\end{IEEEbiography}

\begin{IEEEbiography}[{\includegraphics[width=1in,height=1.25in,clip,keepaspectratio]{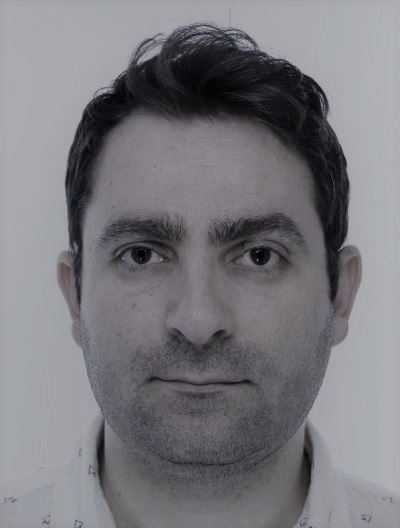}}]{Gholamreza Anbarjafari} heads the intelligent computer vision (iCV) research lab in the Institute of Technology at the University of Tartu. He is an IEEE Senior member and the Vice Chair of the Signal Processing / Circuits and Systems / Solid-State Circuits Joint Societies Chapter of the IEEE Estonian section. He received the Estonian Research Council Grant (PUT638) and the Scientific and Technological Research Council of Turkey (Proje 1001 - 116E097) in 2015 and 2017, respectively. He has been involved in many international industrial projects. He is expert in computer vision, human-robot interaction, graphical models and artificial intelligence. He is an associated editor of several journals such as SIVP and JIVP and have been lead guest editor of several special issues on human behaviour analysis. He has supervised over 10 MSc students and 7 PhD students. He has published over 100 scientific works. He has been in the organizing committee and technical committee of conferences such as ICOSST, ICGIP, SIU, SampTA, FG and ICPR. He is organizing a challenge and a workshop on in FG17, CVPR17, and ICCV17.
\end{IEEEbiography}


%\vfill


\end{document}